\documentclass[10pt,journal,compsoc]{IEEEtran}
\usepackage{amsmath,amsfonts}
\usepackage{algorithmic}
\usepackage{algorithm}
\usepackage{array}
\usepackage[caption=false,font=normalsize,labelfont=sf,textfont=sf]{subfig}
\usepackage{textcomp}
\usepackage{stfloats}
\usepackage{url}
\usepackage{verbatim}
\usepackage{graphicx}
\usepackage{cite}
\usepackage[table,xcdraw]{xcolor}
\usepackage[colorlinks,bookmarksopen,bookmarksnumbered,citecolor=blue, linkcolor=blue, urlcolor=black]{hyperref}
\usepackage{mathrsfs}

\DeclareMathOperator\mask{Mask}

\graphicspath{{figures/}}
\begin{document}

\title{Unification of Closed-Open Industrial Detection Scenarios: New Large-Scale Benchmarks, Challenges and Baselines}

\author{Zekai Zhang$^{\dag}$,~\IEEEmembership{Student Member,~IEEE,} Jinglin Zhang*$^{\dag}$, Qinghui Chen, Gang Li, Da Chen, Shuainan Jing, He Wang, Dagang Li, Cong Liu, Cong Bai, Shengyong Chen, ~\IEEEmembership{Senior Member,~IEEE}
\thanks{
Zekai Zhang, Jinglin Zhang and Qinghui Chen are with the School of Control Science and Engineering, Shandong University, Jinan 250061, China. \\
Da Chen is with CEREMADE, University Paris Dauphine, PSL Research University, CNRS, UMR 7534, 75775 Paris, France. \\
Gang Li, Shuainan Jing, and He Wang are with the Shandong Computer Science Center, Qilu University of Technology, Jinan, China.\\
Cong Bai is with College of Computer Science and Technology, Zhejiang University of Technology, Hangzhou 310023, China.\\
Cong Liu is with the NOVA Information Management School, Nova University of Lisbon, 1070-312 Lisbon, Portugal.\\
Dagang Li is with the School of Computer Science and Engineering, Macau University of Science and Technology, Macau SAR, Macau, China.\\
Shengyong Chen is with the School of Computer Sciences and Engineering, Tianjin University of Technology, Tianjin 300384, China.\\
$^{\dag}$ Zekai Zhang and Jinglin Zhang contributed equally to this work.\\
*corresponding author is Jinglin Zhang (e-mail: jinglin.zhang@sdu.edu.cn).
}
}



\markboth{Unification of Closed-Open Industrial detection scenarios: New Large-Scale Benchmarks, Challenges and Baselines}%
{Shell \MakeLowercase{\textit{et al.}}: A Sample Article Using IEEEtran.cls for IEEE Journals}

\maketitle

\begin{abstract}
Large-scale Visual-Language Models (LVLMs) have achieved remarkable success in natural visual tasks, yet their application to industrial defect detection remains challenging due to two fundamental limitations: (i) the scarcity of large-scale industrial datasets that cover diverse defect categories across multiple domains, and (ii) the reliance on manual prompts (points, boxes, masks) that introduce subjective noise and lack text-visual interaction for fine-grained understanding. To address these challenges, we introduce a Large-Scale Multi-Modal Industrial Open-Closed benchmark (MMIOC-1M) containing over one million samples across $14$ super-categories, $29$ industrial scenes, and $351$ defect subcategories. To our knowledge, MMIOC-1M is the first unified largest benchmark supporting both open-vocabulary and closed-set industrial detection, providing valuable pre-training data for LVLMs in industrial scenarios. Furthermore, we propose a Refined Text-Visual Prompt Network (RTVPNet) that incorporates three key innovations: (1) an expert-assisted domain projection mechanism that enables rapid adaptation of general vision models to industrial domains, (2) an energy-based sparse sampling strategy that automatically generates refined visual prompts without manual intervention, and (3) a bidirectional text-visual interaction module that enhances cross-modal semantic alignment and understanding. Extensive experiments demonstrate that RTVPNet achieves state-of-the-art performance on MMIOC-1M, LVIS, and COCO benchmarks while maintaining computational efficiency. The dataset and code are available at \url{https://github.com/hellozzk/MMIO}.

\end{abstract}

\begin{IEEEkeywords}
Industrial open detection, Large-scale industrial benchmark, Visual-Language Model.
\end{IEEEkeywords}

\section{Introduction}
\IEEEPARstart{P}{roduct} defect detection plays a crucial role in the manufacturing industry and is of great importance in improving product quality and production efficiency. Expert models \cite{wang2022yolov7,2021YOLOX,li2022yolov6,YOLOv8,lite-yolo} in industrial scenarios usually use single-modal data from a single field and strictly follow class-visible methods, which limits the ability of model to process multi-scene data and generalize to open scenarios. Recently, the development of Large-scale Visual-Language Models (LVLMs) \cite{sam,fastsam,fastersam,llava,liu2023groundingdino} has shown powerful interactive and strong generalization capabilities in remote sensing, medicine, and other fields. The uniqueness of these methods lies in the design of human-computer interactive prompts, which allows segmentation based on user-supplied point, line, and box prompts. 

\begin{figure}[!t]
\centering
\includegraphics[width=3.3in]{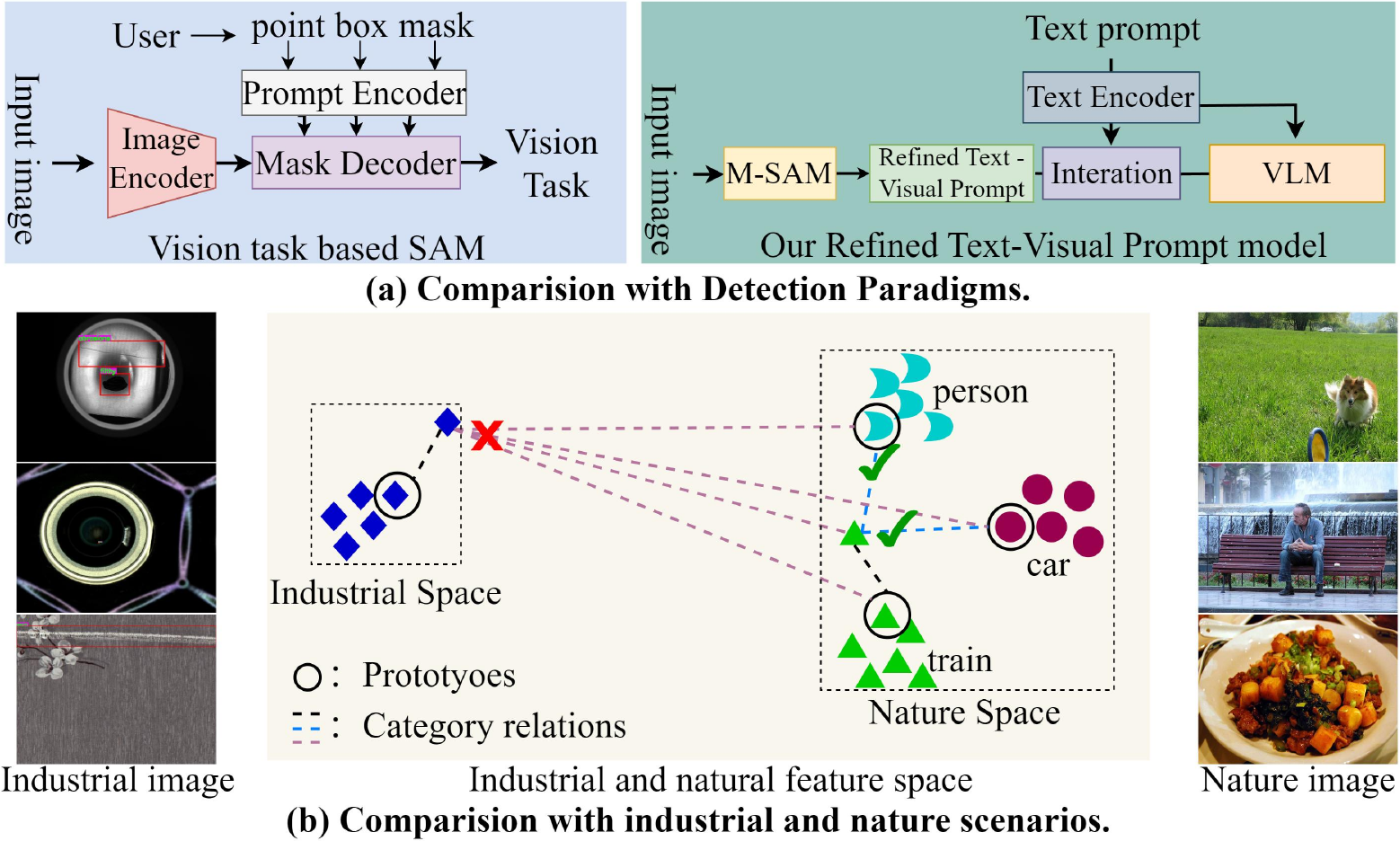}
\caption{(a) Comparison between traditional prompting methods and our method. Our method solves the subjectivity of traditional manual prompts and introduces text to further refine the semantics.(b) Industrial scenes are very different from natural scenes. Models trained in natural scenes are difficult to generalize in industrial scenes. }
\label{Challenge}
\end{figure}

However, there are many significant challenges in applying LVLM's \cite{1,2,3,4,5,6,7,8,9,10,11,12,13,zhang2026novel,zhang2023idd,zhang2025zero,zhang2024representation,zhang2026unification} pre-training-prompt paradigm in industrial scenes. As shown in Fig.~\ref{Challenge} (b), there are significant domain differences between industrial and natural scenarios in the feature space. Simply transferring knowledge from natural scenarios to the industrial defect detection cannot eliminate the significant differences in the fields, so fine-tuning is required for a large amount of domain professional data. However, the existing industrial detection data are all distributed in a single field, and it is impossible to find a unified multi-domain generalized industrial scenario data set. As shown in Fig.~\ref{Challenge} (a), existing LVMLs \cite{sam,fastsam,fastersam,llava} rely on manual work (point, box, mask) to segment the object when processing complex scenes. Faced with the problem that industrial scenes contain complex noise, the user's familiarity can significantly affect the effect of specific prompts and introduce irrelevant or noisy pixels. In addition, most current LVLMs \cite{sam,fastsam,fastersam,llava,liu2023groundingdino} ignore the interaction of visual-text prompts and lack a deeper understanding of industrial scenarios.

The main challenge to applying LVLMs in industrial scenarios is the lack of large-scale data in industrial detection \cite{IndustrialTextileDataset,2020Deep,shouji}, and it is impossible to find a generalized multi-domain unified industrial scene benchmark. To solve the above problem, we created a Large-Scale Multi-Modal Industrial Open-Closed benchmark called MMIOC-1M. MMIOC-1M provides multi-modal visual-text annotations for each category. MMIOC-1M consists of more than 1M+ samples converted from 31 different industrial defect fields. MMIOC-1M is designed for the unique feature distribution in open-closed industrial detection, effectively alleviating the lack of expertise in the industrial domain of LVLMs. To our knowledge, MMIOC-1M is the first large-scale open-closed benchmark for industrial defect detection, and MMIOC-1M can catalyze the development of LVLMs in industrial openness.

In order to solve problems on human prompt, some methods \cite{matcher,personalize} combine semantic models \cite{resnext,resnet} to got pseudo masks of objects. CPT \cite{yao2024cpt} and ReCLIP \cite{subramanian2022reclip} used a visual prompt to establish relationships between instances. Hu et al. \cite{hu2023efficiently} designed a sampling strategy to extract a pseudo template as prompts for SAM. CoCoOp \cite{yan2023cocoopter} turned the image-generated prompt into a conditional input and dynamically combines it with the language prompt. These methods ignore false positives in pseudo masks and rely on human hyperparameter sensitivity. Therefore, they heavily depend on the quality of pseudo masks and have poor generalization ability. In addition, open vocabulary models such as GroundingDino \cite{liu2023groundingdino} and YOLO-World \cite{yoloworld} propose using single-text prompts to strengthen features. However, these methods lack fine-grained image feature prompts. Unlike natural scenes, open industrial scenes present unique challenges. Due to large amounts of noise from invisible categories, it is difficult to maintain robustness in high-noise scenes by simply relying on visual or text prompts. To address the above problems, we propose a Refined Text-Visual Prompt Net (RTVPNet), which improves the open-detection capability of VLMs in industrial scenarios. Based on Mobile-SAM \cite{fastersam} in natural scenes, RTVPNet further enhances its generalization ability in industrial scenes. RTVPNet introduces an expert assistance mechanism based on Mobile-SAM to automatically generate coarse-grained segmentation features and encode these features into a low-dimensional space. In view of the uniqueness of industrial images, we perform energy activation on the segmentation features and extract the uncertainty score of the object. Then, we design a sparse modeling sample selection strategy to extract semantic clues from the enhanced features through the uncertainty score to obtain a Refined Visual Prompt. Finally, the Refined Visual Prompt interacts with the Text Prompt to generate a prompt embedding for a semantically specific object. Building on the inherent capabilities of Mobile-SAM, RTVPNet promotes the capability of the model in understanding and generalizing, especially in industrial open-scenes.

Several experiments on open and closed scenes on MMIOC-1M, LVIS \cite{LVIS}, and COCO \cite{COCO} demonstrate the value of MMIOC-1M and the effectiveness of RTVPNet. Parts of this paper were originally published in AAAI 2025 \cite{Zero-shotLearning}. We extend our previous work in several valuable ways: 1) Compared to MMIO-80K, MMIOC-1M supports defect detection in both open and closed scenarios. MMIOC-1M contains more than 1M samples and 31 industrial scenarios, thereby promoting the development of Large-Scale Multi-Modal Industrial benchmarks. 2) Compared to the previous RTVP version \cite{Zero-shotLearning}, we have newly designed text-visual bidirectional interaction, domain transfer, and energy-based visual prompt optimization methods and added Visual Grounding, Object Detection, and Visual Question Answering tasks. RTVPNet can provide a more detailed and less noisy text-visual prompt than RTVP. 3) We have also added new experiments and more detailed analyzes to prove the advantages of our method.

In summary, our contributions are three-folder:
\begin{itemize}
	\item \textbf{MMIOC-1M Benchmark}. We introduce the first large-scale multi-modal benchmark for unified industrial open-closed detection. MMIOC-1M contains over one million samples across $14$ super-categories, $29$ industrial scenes, and $351$ defect subcategories, supporting multiple downstream tasks including visual grounding, object detection, and visual question answering. This benchmark fills a critical gap in industrial LVLM research by providing comprehensive multi-domain coverage and standardized evaluation protocols for both open-vocabulary and closed-set scenarios.
	\item \textbf{RTVPNet Baseline}. We propose a Refined Text-Visual Prompt Network tailored for industrial detection, featuring three key innovations: (1) expert-assisted domain projection that enables rapid adaptation from natural to industrial domains, (2) energy-based sparse sampling that automatically generates refined visual prompts without manual intervention, and (3) bidirectional text-visual interaction that enhances cross-modal semantic alignment. Experiments on MMIOC-1M, LVIS, and COCO demonstrate that RTVPNet achieves state-of-the-art performance while maintaining computational efficiency.
	\item \textbf{Evaluation}. We establish standardized protocols for industrial open-closed detection and conduct extensive experiments across MMIOC-1M, LVIS, and COCO. Results show the value of MMIOC-1M as a challenging benchmark and validate the effectiveness of RTVPNet against state-of-the-art methods.
\end{itemize}

This paper is organized as follows. Section~\ref{sec_ReWork} introduces the background knowledge of industrial multi-modal datasets and VLMs. In Section~\ref{sec_Benchmark}, the construction process and the analysis of MMIOC-1M are proposed and detailed. Section~\ref{sec_RTVPN} introduces the architectural design of RTVPNet. Section~\ref{sec_Exp} shows the experimental results and conclusions. The discussion and conclusion are presented in Sections~\ref{sec_Discussion} and~\ref{sec_Conclusion}, respectively.

\section{Related Work}
\label{sec_ReWork}
\begin{table*}[t]
\caption{Comparison of MMIOC-1M with large-scale industrial defect dataset. Gen. stands for generate and Misc. stands for synthesize.}
\fontsize{10}{12}\selectfont\rmfamily
\centering
\resizebox{1.5\columnwidth}{!}{

\begin{tabular}{lllllll}
\begin{tabular}{ccccccc}
\hline
Dataset                    & Classes & Number    & Modal                 & Type    & Year & scene category   \\\hline
MMAD \cite{jiang2024mmad}                      & 244     & 8,366     & RGB,Text              & Misc.& 2024 & 38 (Closed)      \\\hline
Defect Spectrum \cite{Defect_spectrum}           & 125     & 5,438     & RGB,Text              & Gen. & 2024 & 14 (Closed)      \\\hline
VISION \cite{bai2023vision}                    & 44      & 18,000    & RGB                   & Misc. & 2023 & 14 (Closed)      \\\hline
PKU-GoodsAD \cite{zhang2024pkugood}               & 12      & 6,124     & RGB                   & Commodity & 2023 & 6 (Closed)       \\\hline
MVTec AD \cite{bergmann2019mvtec}                  & 73      & 5,354     & RGB                   & Misc. & 2019 & 15 (Closed)      \\\hline
VisA \cite{Visa}                      & 78      & 10,821    & RGB                   & Electronic   & 2022 & 12 (Closed)      \\\hline
Real-IAD \cite{wang2024real}                  & 8       & 150,000   & RGB                   & Material & 2024 & 30 (Closed)      \\\hline
MulSen-AD \cite{li2024mulsen}                 & 14      & 2,035     & RGB,3D,Infrared & Material & 2024 & 15 (Closed)      \\\hline
3CAD \cite{yang20253cad}                      & 125     & 27,039    & RGB                   & Electronic & 2025 & 8 (Closed)       \\\hline
Industrial Textile Dataset \cite{IndustrialTextileDataset}  & 10      & 6,000     & RGB                   & Textile  & 2023 & 1 (Closed)       \\\hline
Ind \cite{0Pixel}                       & 30      & 600,000 & RGB                   & Misc. & 2023 & 11 (Closed)      \\\hline
BeanTech \cite{2021VT}                  & 3       & 2,830      & RGB                   & Misc.   & 2021 & 3 (Closed)       \\\hline
\rowcolor[HTML]{EFEFEF} 
MMIO-80K (Previous work) \cite{Zero-shotLearning}                  & 100       & 21,836      & RGB,Text                   & Misc.   & 2024 & 18 (Open-Closed)       \\\hline
\rowcolor[HTML]{E6E0E0} 
MMIOC-1M                   & 351     & 1,000,000 & RGB,Text              & Misc. & 2025 & 29 (Open-Closed)\\\hline
\end{tabular}
\end{tabular}}
\label{Datasetcomp}
\end{table*}

\subsection{Industrial Datasets}
Over the years, the scale of industrial defect detection-related datasets has grown steadily. For multi-domain industrial defect data, the Defect Spectrum \cite{Defect_spectrum} generates images and pixel-level defect labels using a very small amount of industrial defect data, which contains 5,438 defect samples covering 125 types of defects. VISION \cite{bai2023vision} consists of 18,000 images from 44 defect categories. MVTec AD \cite{bergmann2019mvtec} includes 5,354 images of 15 categories of anomaly segmentation. Compared with these three datasets, MMAD \cite{jiang2024mmad} builds a larger multi-modal large language model evaluation benchmark, which contains 8,366 industrial images covering 38 categories of industrial products and 244 defect types, annotated with question-answer pairs in JSON or CSV format. However, these datasets are insufficient to cover multiple industrial fields more comprehensively and in large-scale samples. Although these datasets contain more than 200 categories related to industrial defects, they are different from MMIOC-1M (including 14 supercategories, 29 scenes, and 351 subcategories). On the one hand, general industrial defect detection datasets are difficult to cover a wide range of industrial categories (metallurgy, automobile manufacturing, precision electronics, textiles, daily necessities, basic materials processing, etc.). For example, VisA \cite{Visa} and 3CAD \cite{yang20253cad} mainly cover 3C electronic datection, PKU-GoodsAD \cite{zhang2024pkugood} mainly covers packaging datection, and Real-IAD \cite{wang2024real} and MulSen-AD \cite{li2024mulsen} mainly cover material datection. However, although datasets such as VISION \cite{bai2023vision}, MVTec AD \cite{bergmann2019mvtec}, and MMAD \cite{jiang2024mmad} cover more than five industrial categories, their sample numbers are too small, and they are all tasks in closed scenes, which makes it difficult to expand LVLM to open scenes. 

Considering the important role of large-scale datasets in visual recognition algorithms, especially in the training of LVLMs, we construct a Large-Scale multi-modal Industrial Open-closed scene benchmark MMIOC-1M with more comprehensive category coverage and a larger number of images. In Table \ref{Datasetcomp}, we give the statistics of existing industrial detection datasets and MMIOC-1M. MMIOC-1M exceeds existing datasets in terms of scenes, categories, and sample numbers. To our knowledge, MMIOC-1M is the first open-closed scene unified benchmark proposed in the field of industrial datection. MMIOC-1M can catalyze the development of LVLMs in the open industrial domain.

\subsection{Application of Vision-Language Models}
In recent years, pre-trained Large Language Models, such as GPT-4 \cite{gpt4}, Llava \cite{llava}, etc., have shown strong representation learning capabilities in natural language processing. Subsequently, pre-trained visual-language models such as CLIP \cite{clip}, BLIP-2 \cite{li2023blip}, GroundingDino \cite{liu2023groundingdino} etc., have been extended to computer vision. Currently, there are two methods to apply large pre-trained models. One is to use the segmentation results of large pre-trained models as prior information to assist downstream tasks, which requires additional intermediate layer fine-tuning of the model. For example, Ahmadi et al. \cite{ahmadi2023application} used  SAM segmentation results as priors in crack and other defect detection. Wu et al. \cite{wu2023medicalsam} inserted the adapter module into SAM for medical image segmentation tasks. Another method uses prompts to guide the pre-trained model transfer to the object domain. For example, Xu et al. \cite{xu2023eviprompt} proposed an untrained evidence prompt generation method, incorporating prior human information into prompts. Zhang et al. \cite{jie2023adaptershadow} proposed a shadow detection network to generate dense point prompts. The above two methods easily consume computing resources and cannot guarantee the training effect of the domain layer. In addition, the visual prompts are not refined and do not consider the importance of text prompts. Unlike earlier work, RTVPNet specifically focuses on defect detection in open industrial scenarios. In order to adapt LVLM to industrial tasks, we introduce an expert model to assist Mobile-SAM in generating industrial prior information. In terms of prompting, we use refined text-visual prompts to provide richer industrial semantic information.

\subsection{Prompt for Representation Learning}
The prompt technology originated from NLP. The prompt was subsequently used to guide representation learning in an open scenario. However, prompts often rely on artificial features, leading to user burden and noise introduction. Recently, automated prompt training methods have been widely used in representation learning. For example, AutoPrompt \cite{shin2020autoprompt} used gradient-based methods to automatically generate prompt templates. With the development of large models, fine-grained visual prompts are widely used in open scenarios. For example, CPT \cite{yao2024cpt} introduces colored object boxes as markers in the image. However, the prompts contain a lot of noise. To solve this problem, FGVP \cite{yang2024fine}, VRP-SAM \cite{sun2024vrp} etc., proposed refined visual prompts. However, these methods are based on the adaptability of SAM and cannot iteratively optimize the quality of the prompts. In addition, these methods ignore the role of text prompts in representation learning. Therefore, some methods based on text prompts have been widely proposed. For example, GroundingDino \cite{liu2023groundingdino} detects semantically related objects through text prompts, and YOLO-UniOW \cite{liu2024yoloUNI} performs open-world and vocabulary tasks through text prompts. However, existing open-vocabulary methods usually use large-scale text prompts and image pre-training to match image-related regions but ignore the role of refined visual prompts for correct text-image matching. Unlike the above methods, we focus on the promotion and adaptation of LVLMs to the industrial field and propose an optimized text-visual interaction prompt strategy.

\section{Multi-Modal Industrial Open-Closed Benchmark}
\label{sec_Benchmark}

\subsection{Benchmark Construction}
We create a unified Large-Scale multi-modal Industrial Open-Closed-scene benchmark called MMIOC-1M. MMIOC-1M consists of more than 1M image-text sample pairs converted from 31 different industrial scenarios, including 351 product defects in 17 major industrial categories. MMIOC-1M is the first Large-Scale Multi-Modal pre-training benchmark for industrial detection, providing valuable training data for large models in future industrial scenarios. We have completed the construction of MMIOC-1M by extensively collecting defect diagrams of 47 product companies and organizations, and it has been authorized. The construction process is divided into three stages: the construction of category attribute of defects, image screening and calibration, and the division of open-closed subsets.

\subsubsection{Attribute of Defect Categories} 
\begin{figure*}[!t]
\centering
\includegraphics[width=6.5in]{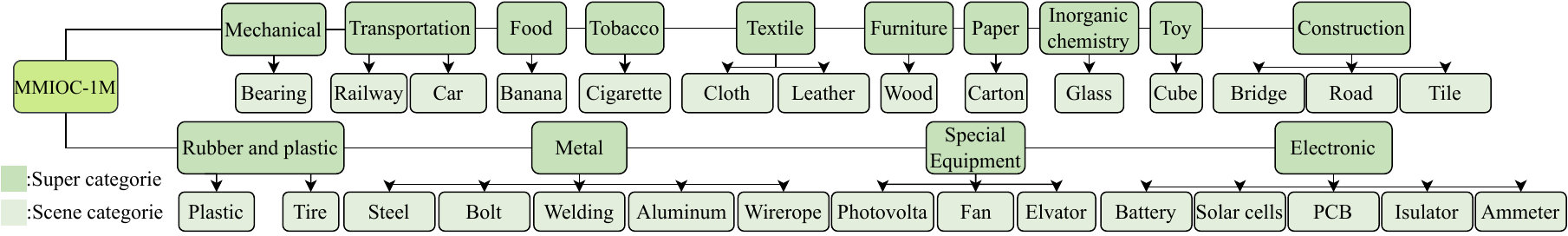}
\caption{MMIOC-1M category attribute analysis. MMIOC-1M contains 14 supercategories, 29 scene categories, and 351 subcategories. To our knowledge, MMIOC-1M is the largest unified defect benchmark for industrial open and closed scenarios.}
\label{category}
\end{figure*}

\begin{figure}[!t]
\centering
\includegraphics[width=3.2in]{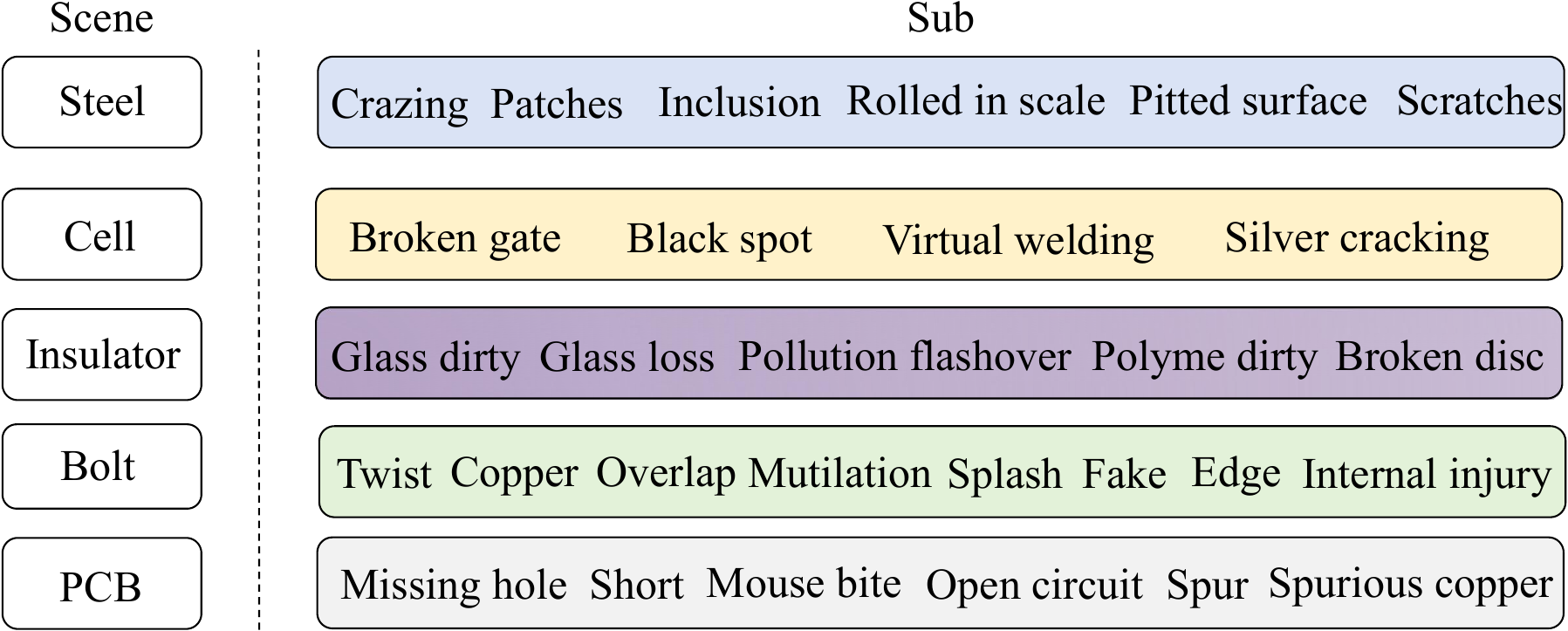}
\caption{Subcategories of some scene categories in MMIOC-1M. Each scene in MMIOC-1M contains multiple categories with similar semantics, which brings challenges to the classification of LVLMs.}
\label{categoryexample}
\end{figure}

Since there are many category samples in MMIOC-1M, a detailed attribute is needed to avoid semantic confusion among categories. We adopt a top-down approach. Based on the characteristics of the data, we first define 14 major industrial categories to construct supercategories and then subdivide 351 subcategories of 29 scenarios under each supercategory. Considering that it is time-consuming and difficult to add high-level attributes to the categories of each sub-dataset manually, we conduct the following iterative process: first, all categories of each sub-dataset are screened by experts to obtain the category names and few image features, and then GPT-4V is used to perform semantic retrieval of images and category names in the industrial categories specified by the United Nations to obtain the super categories to which subcategory belongs. Finally, we verify and aggregate the retrieved supercategories through strict manual verification. For example, the sub-dataset category set \{Damaged clothing, damaged zippers, holes in leather goods\} is aggregated into the supercategories "Textile". For the scene category, we identify 29 scenes for the product categories of 47 sub-datasets, which contain a total of 351 subcategories. Fig.~\ref{category} shows the attribute relationship between supercategories and scene categories, which shows that MMIOC-1M contains rich attributes and can cover most industrial categories. Fig.~\ref{categoryexample} shows some subcategories under the scene category.

\begin{figure}[!t]
\centering
\includegraphics[width=3.2in]{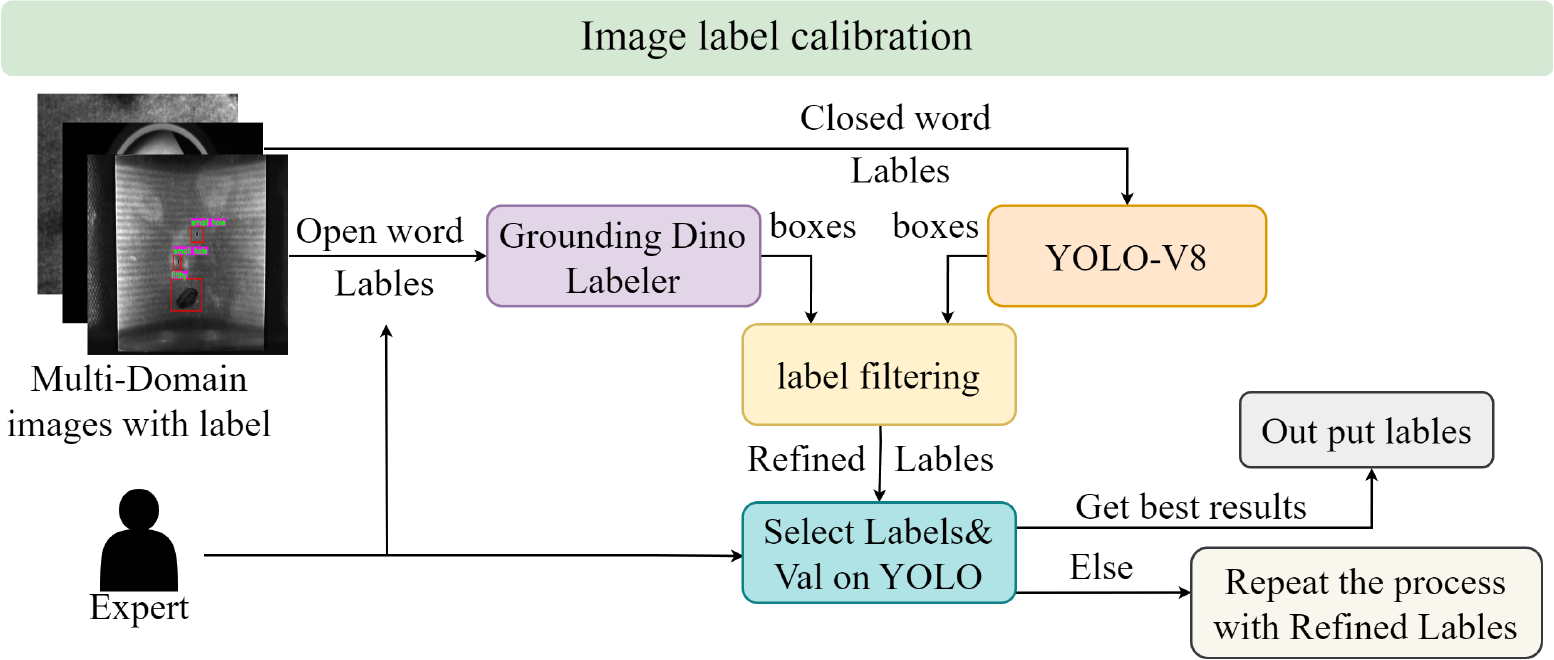}
\caption{Label calibration process. According to the category properties we have defined, jointly calibrating label categories through detectors of open scenes and closed scenes can save manpower.}
\label{label}
\end{figure}

\subsubsection{Image Screening and Calibration} 

MMIOC-1M contains more than 1,000,000 defect samples. Although some samples have been equipped with ground truths and classification labels, the labels of most samples are still incorrect. Ideally, experts should sort the images and annotate them one by one. However, in the face of a large number of samples, expert annotation is time-consuming and laborious. Therefore, we only selected 15\% of the samples for fine-grained annotation by experts, and the remaining samples are automatically annotated by the designed adversarial annotation method. As shown in Fig.~\ref{label}, we introduce GroundingDino \cite{liu2023groundingdino} and YOLOV8 \cite{YOLOv8} to extract the semantic vector of each defect image and generate predicted boxes. For open scenes, we use GroundingDino to train 20\% of the samples to calculate the similarity score between the semantic vector of the entity in MMIOC-1M and the semantics related to the image. Then, we select the text-image matching pair with the highest similarity using a threshold of 0.5. For closed scenes, we use YOLOV8 to train 20\% of the samples to generate the predicted boxes and categories. The label filter is used to merge labels and remove redundant labels. Finally, experts verified the text-image matching results for 351 categories, corrected some labels, and repeated the above process. Since two models were used for automatic annotation, expert review was an important step, but adversarial automatic annotation still saved a lot of time.
\subsubsection{Annotation quality analysis}
We verify the reliability of the annotations through random manual review, quantitative consistency assessment, scene-level bias analysis, and noise robustness. We randomly sampled 6,000 images from 29 scenes in the automatically annotated images and had three industrial inspection experts re-annotate 351 types of defects. As shown in Table \ref{MMIOCQuality}, the IoU, Cohen’s $k$, IoU Precision, and Recall further demonstrate the precision of automatic annotations. An ANOVA test was performed on the IoU of the 29 industrial scenes, yielding p=0.12$\gg$0.05, which confirms the null hypothesis that “there is no systematic difference in the quality of the annotation in the scenes.” Further injection of 5\% random-label noise into the RTVPNet-S training resulted in a decrease in AP of only 0.006, demonstrating the robustness of the model to annotation errors. Furthermore, the defect visualization in Figure \ref{datasetvis} also shows (e.g., tiny defects such as bananas and PCBs are accurately annotated) that the MMIOC-1M annotation accuracy can support large-scale pre-training and fair evaluation.
\begin{table}[t]
\caption{Annotation quality analysis.}
\fontsize{16}{18}\selectfont\rmfamily
\centering
\resizebox{1\columnwidth}{!}{
\begin{tabular}{lll}
\hline
Metrics                & Value  & Notes                                                                   \\ \hline
Cohen’s $k$ (351 Classes) & 0.972  & \begin{tabular}[c]{@{}l@{}}Consistency of category labels between automatic \\ and expert annotations\end{tabular} \\
IoU$\ge$0.95               & 0.997  & Automatic and expert annotation of box IOU                              \\
Precision (IoU$\ge$0.95)      & 0.998  & Automatic and expert annotation accuracy                                  \\
Recall (IoU$\ge$0.95)         & 0.997  & Automatic and expert annotation recall                                    \\
5\% Tag Noise $\triangle$AP50    & -0.006 & 351 categories average                                                  \\ \hline
\end{tabular}}
\label{MMIOCQuality}
\end{table} 

\subsubsection{Division of Open-Closed Subsets} 

For closed scene data, we divide the 351 defect samples in MMIOC-1M into training and validation sets in a ratio of 7:3. For open scenes, it is necessary to generalize from limited annotated visible categories to invisible categories. In the open scenes task, the high similarity of the semantic embedding space between visible and invisible classes supports the migration of discriminative boundaries \cite{chaudhuri2024relational}. Specifically, semantically similar categories are adjacent in the embedding space, and the same hyperplane can cover the overlapping distribution areas of their visual features through shared interval constraints. The interval constraints of the model on the visible categories can be generalized to semantically similar invisible categories to achieve cross-category generalization. Based on the above principles, we use YOLOv8 to extract the semantic vector of each type of defect and use the cosine distance to calculate the similarity score between each MMIOC-1M image and the category-related semantics. According to the similarity threshold of 0.1, we have 94 visible and 64 invisible classes. To ensure that semantically similar categories are adjacent in the embedding space, we use a description method that combines scene categories and subcategories (e.g., holes in wood) rather than overly detailed descriptions (such as "fragments around edges and corners") to ensure the generalizability of the data.

\subsection{MMIOC-1M Hierarchy, Statistics, and Challenges}

\subsubsection{MMIOC-1M Hierarchy} 

Considering that MMIOC-1M contains 351 different defects of 29 products, we establish a three-level data classification system by establishing 29 product defects and existing industrial major category systems. MMIOC-1M contains 14 supercategories (such as "Food" and "Paper"), and each supercategory corresponds to 1-3 scene categories (such as "Steel" and "Aluminum" in "Metal"). Each scene contains many defects (such as "broken" and "dirty" on the "steel strip"). Fig.~\ref{category} shows the attribute relationship between the supercategories and the category of scenes. Unlike comprehensive general industrial datasets, MMIOC-1M contains rich category attributes and can cover most industrial categories to be used as a benchmark for LVLMs.
\begin{figure*}[t]
\centering
\includegraphics[width=6.5in]{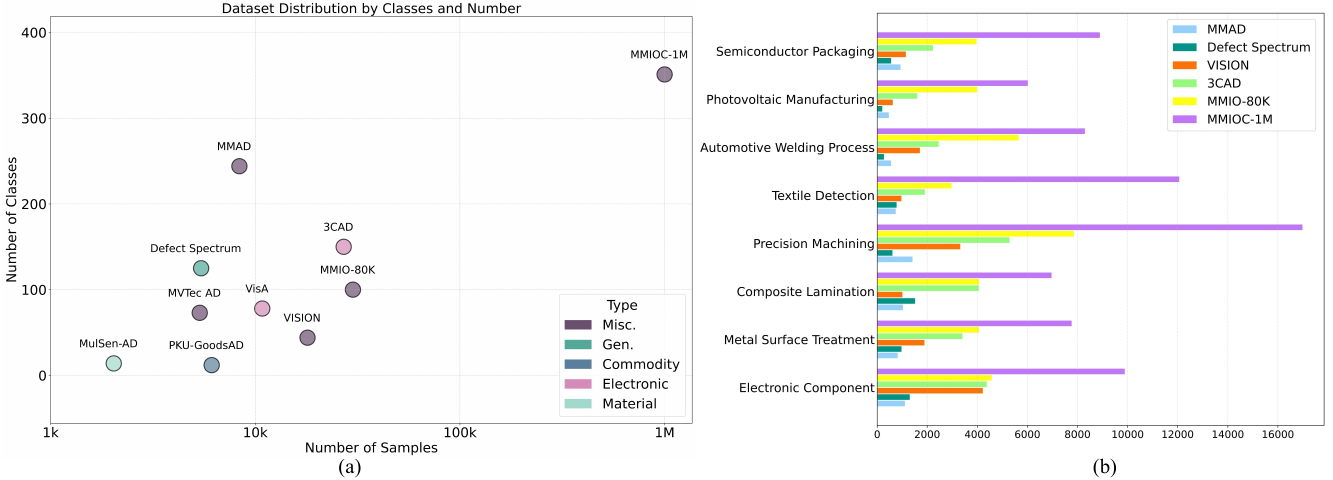}
\caption{Comparison of MMIOC-1M with other industrial datasets. (a) MMIOC-1M has significant advantages in categories and quantity. (b) MMIOC-1M has significant advantages in various industrial defect categories.}
\label{datasetcompare}
\end{figure*}

\subsubsection{Characteristics and Statistics} 

\begin{figure}[t]
\centering
\includegraphics[width=3.2in]{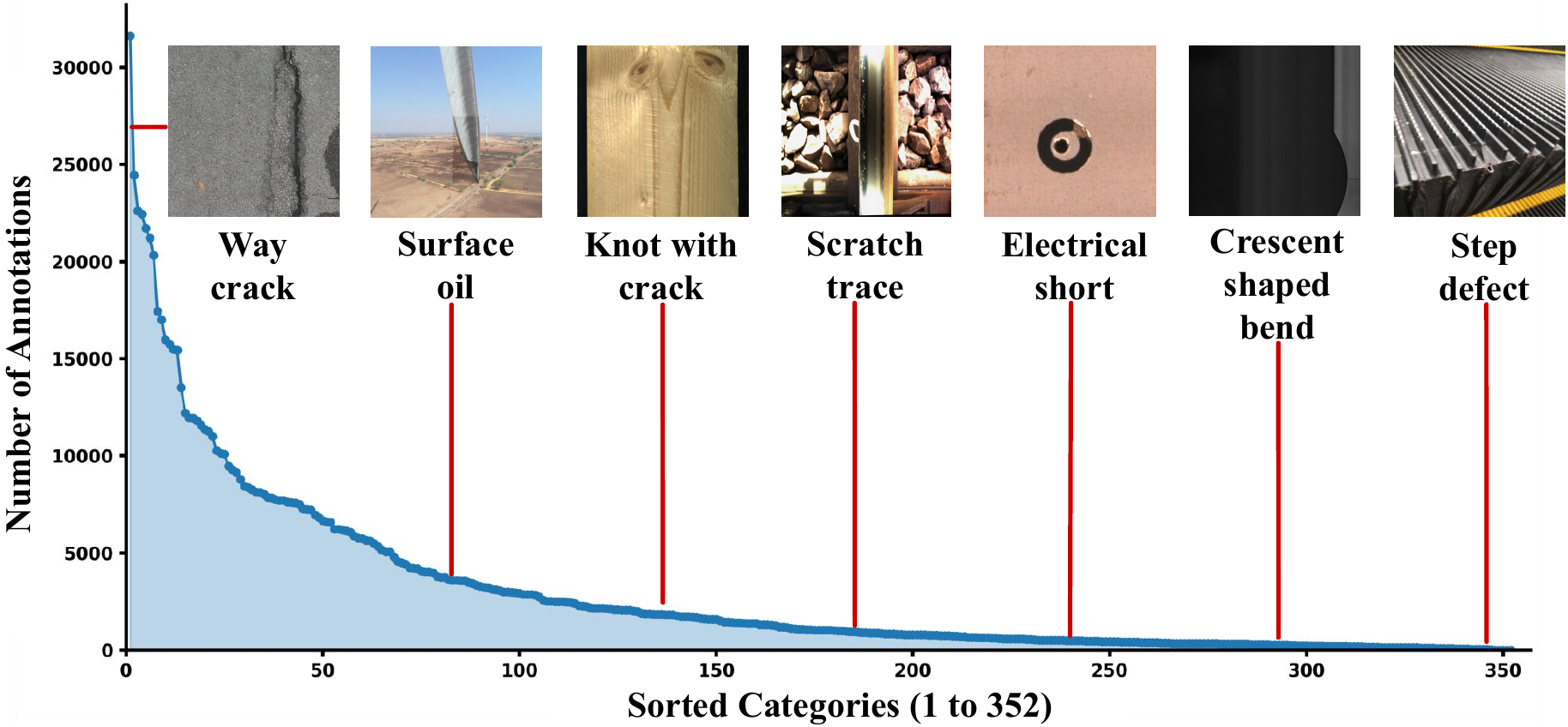}
\caption{The distributions over each category in the MMIOC-1M. It can be observed that MMIOC-1M exhibits characteristics of a long-tail distribution and drastic scale changes.}
\label{statistic}
\end{figure}

\begin{figure}[!t]
\centering
\includegraphics[width=3.2in]{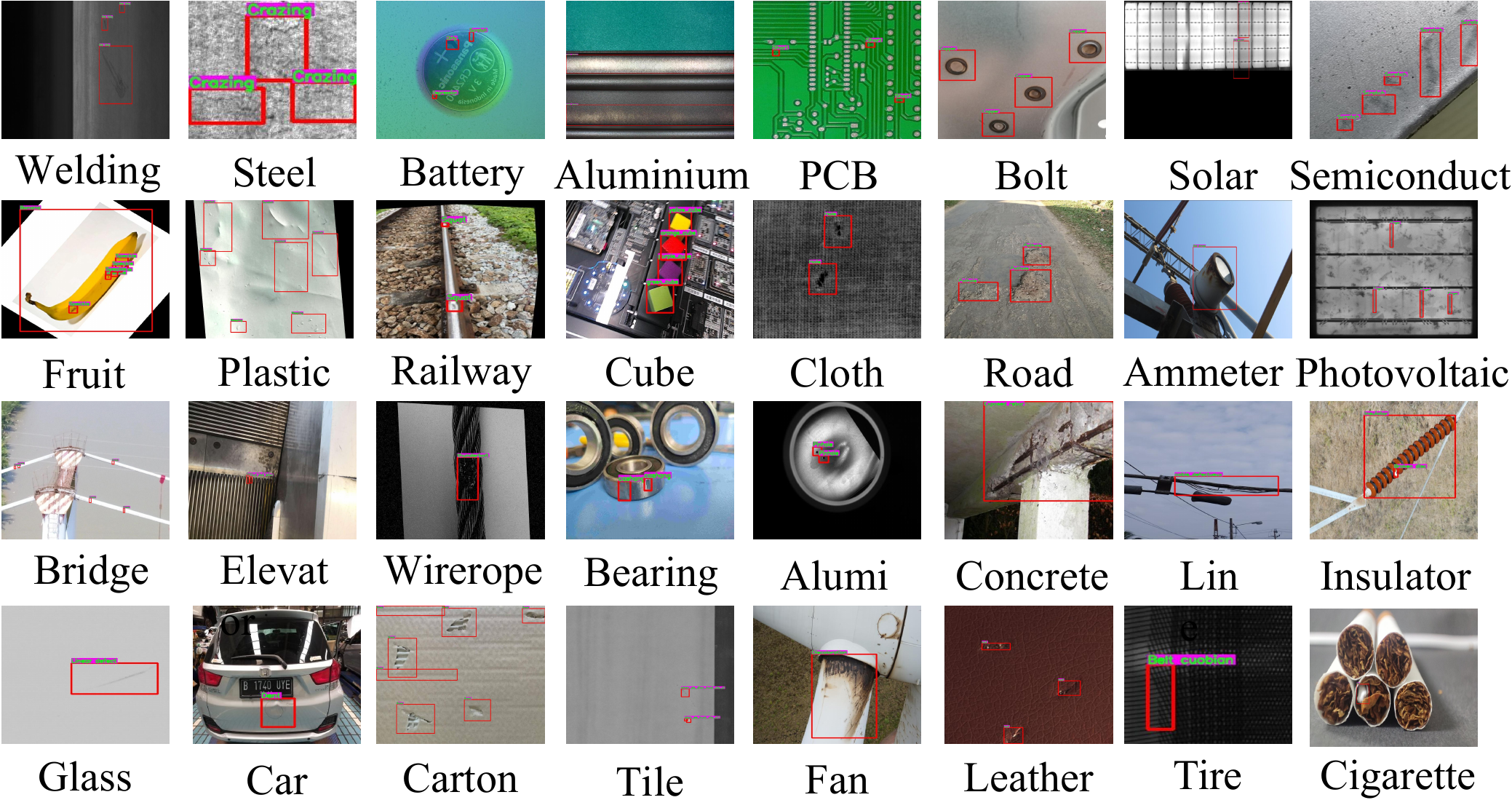}
\caption{Visualization example of MMIOC-1M. MMIOC-1M contains multiple industrial product defects in different scenarios and scales, which poses a great challenge to pre-training existing LVLMs.}
\label{datasetvis}
\end{figure}

Fig.~\ref{statistic} counts the object categories in MMIOC-1M in descending order. The number of images in each category is in the range of [1000, 30000], which is a significant challenge to class imbalance. Fig.~\ref{datasetvis} shows defects in some scenes. The types of product and manufacturing processes in different industrial fields will produce different defects, among which small defects account for the vast majority. In particular, MMIOC-1M stands out with rich attribute annotations, covering a wide range of industrial manufacturing categories, making it particularly suitable for the complex task of industrial open-closed scene. Table \ref{Datasetcomp} compares the comprehensive datasets of standard industrial defects, showing that MMIOC-1M has significant advantages in closed-open scenes and semantic annotation. Fig.~\ref{datasetcompare} shows the scale of MMIOC-1M and compares it with existing industrial datasets. As shown in Fig.~\ref{datasetcompare} (a), the number of categories and samples in MMIOC-1M is an order of magnitude greater than previous industrial datasets. Fig.~\ref{datasetcompare} (b) provides a distribution of several defect categories and compares them with typical datasets. For each category, MMIOC-1M is more significant than the existing datasets. MMIOC-1M is the first defect dataset for industrial open-closed scenarios. In addition, MMIOC-1M specializes in attribute enhancement for representation learning tasks, providing a more challenging and relevant benchmark for representation learning.

\subsubsection{MMIOC-1M Characteristics} 
\begin{figure}[t]
\centering
\includegraphics[width=3.3in]{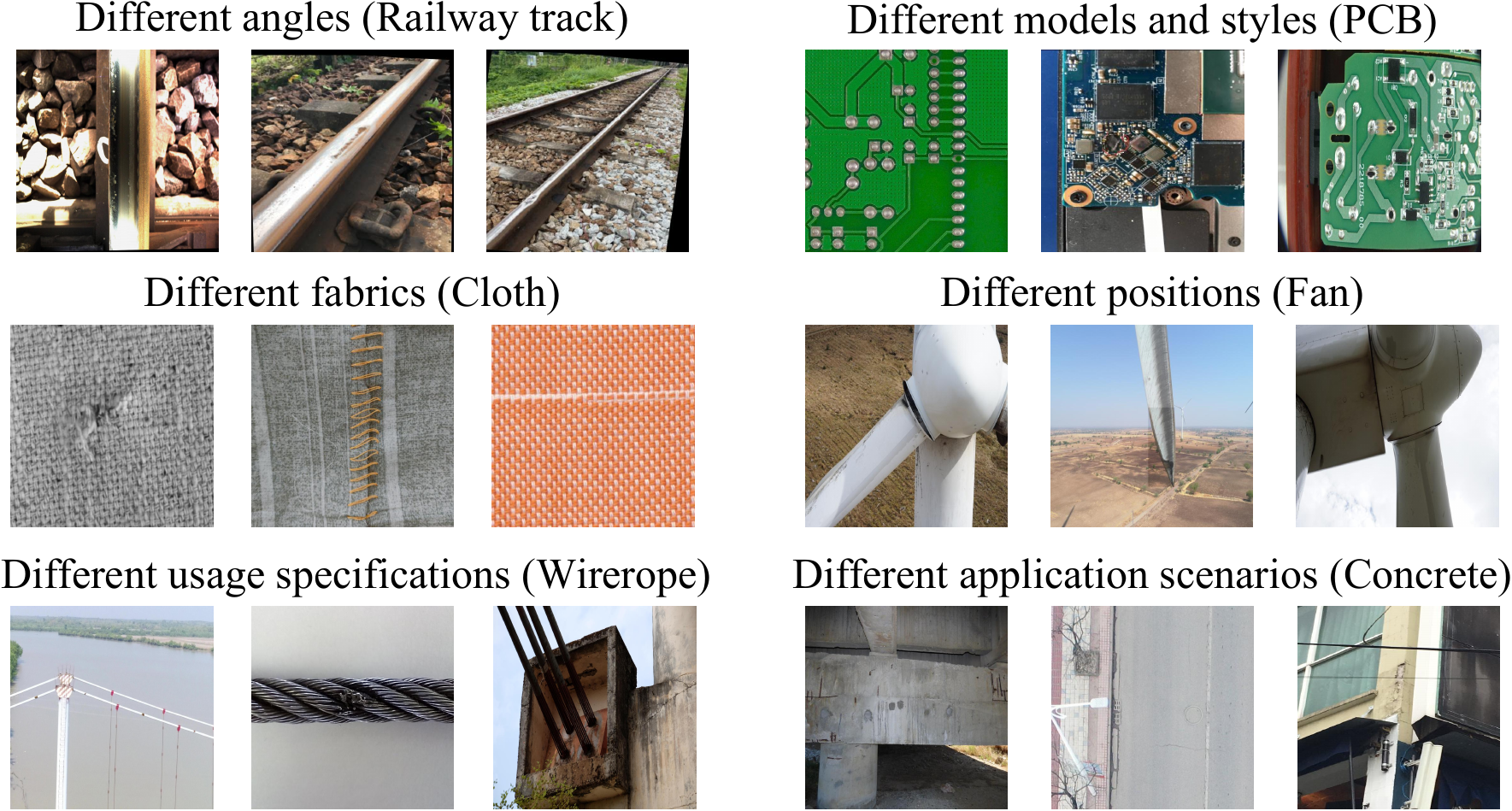}
\caption{Challenging examples in MMIOC-1M. MMIOC-1M has the characteristics of dramatic scale changes, the inter-class differences and out-of-class similarities, as well as extremely rich attributes which lead to difficulties for representation learning.}
\label{differentsamples}
\end{figure}

In addition to the characteristics of open scene detection, more comprehensive coverage of defects, and a more significant number of images, MMIOC-1M also has the following characteristics: First, MMIOC-1M covers a broader range of defect appearances. Fig.~\ref{differentsamples} shows some examples. Noise interference and intra-class correlation make the discriminant information of defects unclear. For example, a PCB contains many similar but different defects, and the defects in cloth have different representations, but belong to the same category. The uniqueness of MMIOC-1M leads to higher intra-class differences, making large-scale pre-training of LVLMs difficult. Second, the defect representation of MMIOC-1M in different industrial scenes varies considerably, resulting in drastic scale changes, for example, between the railway and the PCB in Fig.~\ref{differentsamples}. Finally, there is the long-tail distribution problem. As shown in Fig.~\ref{statistic}, industrial scenes need to seek balanced detection effects in extremely unbalanced category distributions. All of the above factors make MMIOC-1M a new and challenging Large-Scale Multi-Modal industrial benchmark.

\begin{figure*}[t]
\centering
\includegraphics[width=7in]{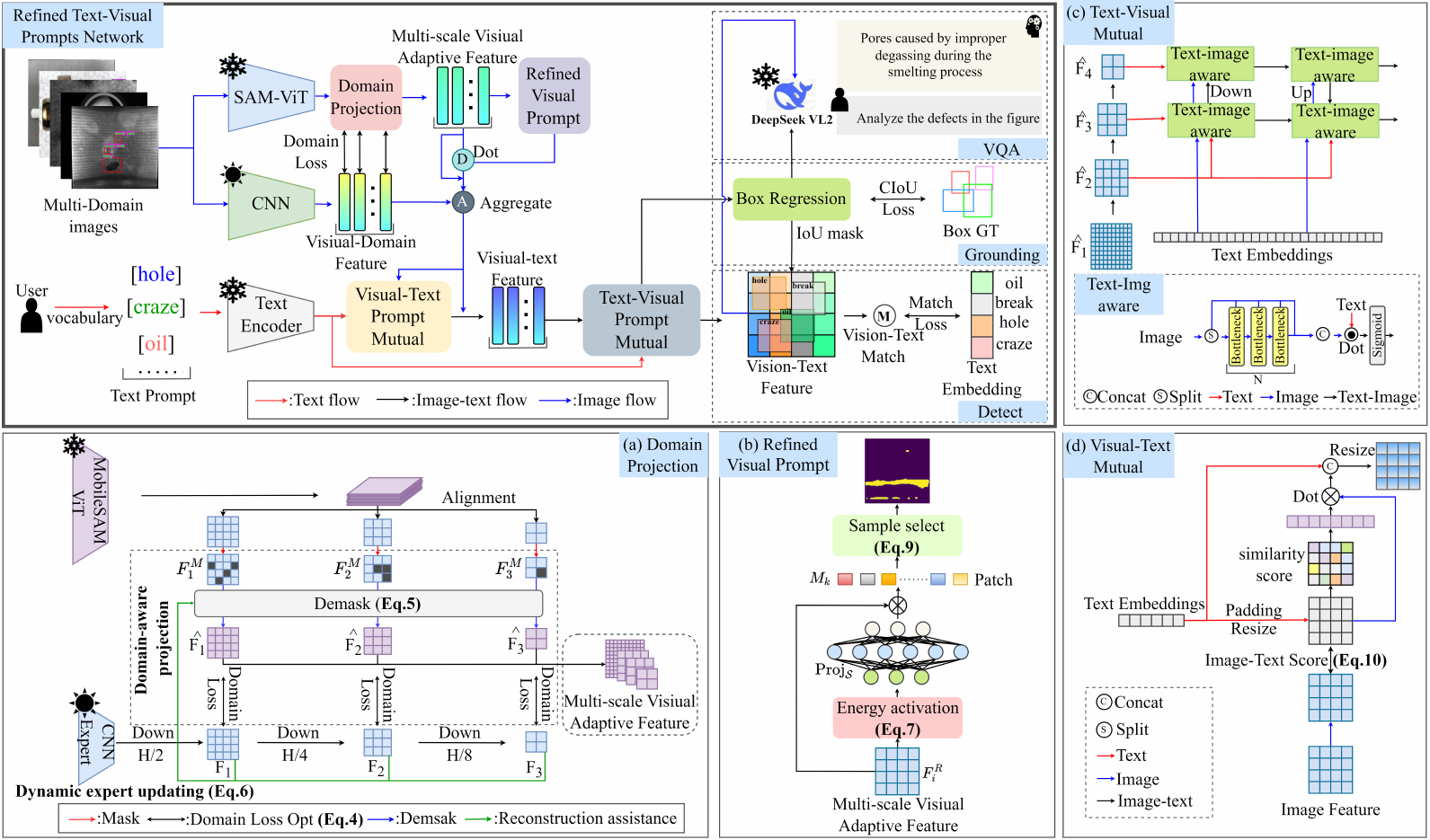}
\caption{RTVPNet framework. RTVPNet contains three basic tasks: Visual Grounding, Vision Question Answering, and Object Detection. (a) Domain Projection includes expert model pre-training, domain-aware projection, and dynamic expert update. The convergence speed of the expert CNN pre-trained in industrial scenarios can provide domain migration for Mobile-SAM quickly. (b) Refined Visual Prompt to establish a sample selection strategy and generate refined visual prompts to promote the understanding of specific tasks in industrial scenarios. (c) Text-Visual Mutual refines visual features in multi-scale feature through text-guided interaction of visual features. (d) Visual-Text Mutual integrates image channels and spatial dimensions into similar regions of text, which helps to align and enhance the semantics of image and text.}
\label{Architect}
\end{figure*}

\section{Refined Text-Visual Prompt Network}
\label{sec_RTVPN}
\subsection{Overview of RTVPNet}
As shown in Figure \ref{Architect}, RTVPNet unifies the three main tasks of visual grounding, question answering, and detection. RTVPNet comprises three core components: Expert-assisted domain projection enables rapid industrial-domain migration of general MobileSAM; Refined Visual Prompt filters high-confidence defects on sparse energy maps to enhance understanding of specific tasks in industrial scenarios; and Bidirectional Text-Visual prompt interaction enables semantic alignment and enhancement between image and text.

\subsection{Problem Formulation}
The goal of the open scene framework is to detect objects that have never appeared in the domain under pre-training with domain-specific text $Y\subseteq \varphi^{N} =\left\{t^{1},...,t^{N} \right\}$ and image $I\subseteq\delta ^{N} =\left\{ I^{1},...,I^{N}\right\}$. We provide a training dataset $D_{s}$ containing image-text pairs of $C_{s}$ visible categories. Let $z_{s}=\left \{1,..., C_{s}\right\}$ and $z_{u}=\left \{1,..., C_{u}\right\}$ are the label sets of visible and invisible categories, respectively. $z_{s}\cap z_{u}=\phi$. $D=D_{s}+D_{u}$ is the image-label space set of visible and invisible classes. Let the text set $Y=y_{s}+y_{u}$. During training, the model extracts the semantic information of $y_{s}$ contained $z_{s}$ and accurately matches $z_{s}$ to the relevant area of image $I$. A test set $D_{t}$ contains $D_{s}$ and $D_{u}$. The goal of the open task is to optimize a model from $D_{s}$ and detect the invisible category $C_{u}$ in $D_{u}$ through the user-defined invisible text prompt $y_{u}$, where $y_{u}$ contains the semantic information of $z_{u}$. For the closed scene, we test the $D_{s}$ category accuracy in $D_{t}$ to measure the effect of the visible class.  

\subsection{Expert-assisted Domain Projection} 
\label{Expert-assisted Domain Projection}
\textbf{Motivation}: Although Mobile-SAM has excellent generalization capabilities, it still faces the challenge of feature distribution shift in industrial scenarios \cite{Segment-Not-Perfect}. Previous solutions can be roughly divided into two categories. One is to insert a learnable layer into SAM to perform domain adaptation \cite{wu2023medicalsam}, and the other is to adjust the prompt of SAM to make it conform to the feature distribution of the target domain \cite{yang2024fine,sun2024vrp}. However, the lack of supervision signals and knowledge understanding capabilities of these two methods leads to suboptimal convergence. Thus we propose an expert collaborative domain projection framework, as shown in Fig. \ref{Architect} (a). It establishes a two-stream feature mechanism from the pre-trained expert model $M_{expert}$ to Mobile-SAM to provide Mobile-SAM with expert knowledge injection in the industrial field.

Specifically, suppose that, the feature distributions of the pre-trained source domain $\mathbb{P}_{p}$ and the industrial domain $\mathbb{P}_{i}$ trained by Mobile-SAM satisfy $\mathbb{P}_{p} \neq \mathbb{P}_{i}$. There exists a projection function $\Phi: \mathbb{P}_{p} \rightarrow \mathbb{P}_{i}$ such that $d_{\mathcal{H}}\left(\Phi\left(\mathbb{P}_{p}\right), \mathbb{P}_{i}\right) \leq \epsilon$. Where $d_{\mathcal{H}}$ is the distribution distance metric under the reproducing kernel Hilbert space. The core of Mobile-SAM's adaptation to industrial scenarios is to find a suitable projection function $\Phi$ to transfer the output feature distribution $\mathbb{P}_{p}$ of Mobile-SAM to the feature distribution of industrial scenarios $\mathbb{P}_{i}$. To achieve this projection, we introduce an expert model $M_{expert}$ pre-trained for industrial scenarios as a supervision source and constructs a two-stream architecture as shown in Fig.~\ref{Architect} (a). Expert-assisted domain adaptation contains three core components: Expert Model pre-training, Domain-aware Projection, and Dynamic Expert Updating.

\subsubsection{Expert Model Pre-training} 

We use the C2F model \cite{YOLOv8} to pre-train visible categories $z_{s}$ in the pre-training dataset $D_{s}$. The purpose of expert model training is to learn the knowledge of industrial scenarios and build a basic feature embedding space $\varepsilon \in \mathrm{R}^{\mathrm{d}}$ that conforms to the distribution of industrial scenarios to minimize category conditional risks. The above process can be expressed as:
\begin{equation}
\min _{\theta_{E}} \mathbb{E}_{(x, y) \sim P_{i}}\left[\mathcal{L}_{\mathrm{cls}}\left(f_{\phi}\left(g_{\theta}(x)\right), y\right)+\lambda_{1}\|\theta\|_{\mathcal{F}}\right],
\end{equation}
where $g_{\theta}: \mathbb{R}^{H \times W \times 3} \rightarrow \mathcal{E}$ represents the expert model with parameter $\theta$, $f_{\phi}: \mathcal{E} \rightarrow \mathbb{R}^{C}$ is the classification head network, $\mathcal{L}_{\mathrm{cls}}$ is the cross entropy error loss function, $x$ and $y$ are the predicted value and the true value respectively, and $\lambda_{1}\|\theta\|_{\mathcal{F}}$ is the regularization term. 

\subsubsection{Domain-aware Projection} 
In order to realize the industrial knowledge transfer of Mobile-SAM, we design a domain-aware projection module $\Phi$. Specifically, given the multi-scale features $F_{i}$ of Mobile-SAM and the multi-scale features $F_{j}$ of $M_{expert}$, we perform multi-scale partitioning on $F_{i}$ and decomposes it into $N=2^{k}\left(k \in \mathbb{N}^{+}\right)$ neighbourhood blocks $\left\{F_{i}^{N}\right\}_{n=1}^{N}$, and the spatial index of each block is $P_{n}^{c}=\left[\left(x_{n}^{\text {start }}, y_{n}^{\text {start }}\right),\left(x_{n}^{\text {end }}, y_{n}^{\text {end }}\right)\right]$ ($x$ and $y$ represent the relative position of the space block). $N$ indexes are randomly selected according to uniform distribution to generate a binary mask. The above process can be defined by:
\begin{align}
\mask_{x, y}^{(i)}=
\begin{cases}
0, & \text{if}~(x, y) \in \bigcup_{n=1}^{N} P_{n}^{c} \\
1, & \text{otherwise}	
\end{cases}
\end{align}
Unlike MAE \cite{he2022mae}, we reconstruct domain expert information rather than the original input. Giving Mobile-SAM more expert context features to participate in the reconstruction is conducive to rapid domain transfer. The above process can be expressed by the following formula:
\begin{equation}
\begin{array}{c}
F_{i}^{M}=\mathcal{P}\left(F_{i}\right) \odot \operatorname{Mask}_{x, y}^{(i)},
\end{array}
\end{equation}
where $\mathcal{P}(\cdot)$ is spatial pooling, and $\odot$ represents element-by-element multiplication. The reconstruction network $\mathcal{G}_{\theta}$ consists of multiple convolutions and deconvolutions so that Mobile-SAM can fully learn the feature projection in industrial scenarios. The reconstruction process uses the features of $M_{expert}$ as conditions for knowledge injection to get coarse-grained prompt features $F_{i}^{R}$. The above process can be expressed by:
\begin{equation}
\begin{split}
F_{i}^{R}&=\mathcal{G}_{\theta}(F_{i}^{M}+F_{i})\\&=\sum_{i=1}^{3} \sigma(\mathbf{U}_{i}^{T} \uparrow*\mathbf{W}_{i}^{T}\downarrow*(F_{i}^{M}+F_{i})),
\end{split}
\end{equation}
where $\mathbf{W}_{i}$, $\mathbf{U}_{i}$ respectively represent the learnable parameter matrices of convolution and deconvolution, and $\sigma$ is the RELU activation function. The reconstructed features of $F_{i}^{R}=\mathcal{G}_{\theta}\left(F_{i}^{M}\right)$ as coarse-grained visual prompts will be further refined. We introduce a multi-scale domain optimization function to optimize feature reconstruction and solve the problem of insufficient supervision information in the existing Mobile-SAM downstream scene tasks. The above process can be expressed by the following formula:
\begin{equation}
\begin{array}{c}
\mathcal{L}_{\text {opt}}=\sum_{i=1}^{3}\left\|\psi\left(F_{i}\right)-\phi\left(F_{i}^{R}\right)\right\|_{2},
\end{array}
\end{equation}
where $\psi(\cdot)$ and $\phi(\cdot)$ are the feature alignment functions used for achieving channel dimension matching.  Note that the optimization will be terminated immediately  once the (nonnegative) value of $\mathcal{L}_{\text {opt}}$ get to be low sufficiently. 

\subsubsection{Dynamic Expert Updating} 

Unlike the traditional freezing strategy, the expert model parameters $\theta_{\text{expert}}$ are continuously updated during training, forming a co-optimization paradigm. The above process can be expressed by the following formula:
\begin{equation}
\theta_{\text{expert}}^{t+1}=\theta_{\text {expert}}^{t}-\gamma \nabla_{\theta_{\text {expert }}} \mathcal{L}_{\text {opt}},
\end{equation}
where $\gamma=2e-3$ is the coupling learning rate. This mechanism ensures that the expert model is continuously refined during the adaptation process, improving the efficiency of knowledge transfer.
\subsection{Refined Visual Prompt}
\label{Refined Visual Prompt}
\textbf{Motivation}: Existing prompt methods rely on pre-detectors or user prompts, and there are problems such as prompt quality affecting performance and prompt optimization being non-differentiable. Furthermore, industrial images often have small defect areas and strong background textures. If a coarse-grained mask is used directly as a visual prompt, a large number of background noise will be introduced, leading to boundary drift. In contrast, we use Mobile-SAM guided by domain expert knowledge to generate coarse-grained visual prompts and establishes a prompt optimization strategy for sparse modeling to obtain more refined visual prompts. The Refined Visual Prompt cannot only introduce object-related feature knowledge enhancement but also iteratively optimize the generation quality.

As shown in Fig.~\ref{Architect} (b), based on the coarse-grained visual prompts $F_{i}^{R}$ obtained by expert-assisted domain adaptation, the sample optimization strategy of sparse modeling quantifies the pixel-level semantic importance by constructing an energy function to obtain the uncertainty score of each pixel position. Based on the feature suppression theory in neuroscience, the energy of each spatial position $(h,w)$ can be described by
\begin{equation}
\begin{array}{c}
\mathscr{E}=\frac{4\left(\sigma\left(F_{h,w}\right)-\mu\right)^{2}}{\left(1+\sigma\left(F_{h,w}\right)-\mu\right)^{2}+3 \delta^{2}}+\epsilon,
\end{array}
\end{equation}
where $\mu$ is the global mean, $\delta^{2}$ is the variance, $\sigma(\cdot)$ is the Sigmoid normalization representing the channel dimension, and $\epsilon$ is an infinitesimal constant to prevent division by zero. The generation process of the uncertainty score $M_{k}$ can be expressed as follows:
\begin{equation}
M_{(k)}=\operatorname{Proj}_{\mathcal{S}}\left[\mathscr{E}\left(F^{R}_{i}\right) \odot F^{R}_{i}\right],
\end{equation}
where $\operatorname{Proj}_{\mathcal{S}}$ is a two-layer MLP used to filter key features. The two-layer MLP will continuously iterate and optimize the prompt. The feature area with the maximum energy will be enhanced. In the traditional uncertainty sampling method, the sample with the highest uncertainty score will be selected. However, samples with large uncertainty scores are usually the local optimal solutions of the features so that the network will fall into the local optimum. To solve this problem, we create a sparse optimization strategy. Specifically, industrial images have significant sparse characteristics, and the sparse optimization strategy selects the most high-frequency areas in the image to reduce the redundancy of irrelevant features. We use patch segmentation with different sizes of neighbourhood  to retrieve the uncertainty score and obtain the high-frequency information in each patch. 

Given a set of coarse-grained prompt features $F_{i}^{R}$, a sparse selection mechanism is used to select high-frequency pixels in each patch of $N_{x,y}^{p}$. Specifically, based on the local sparse prior of industrial images, sparse sample selection is completed by selecting pixels with pixel values greater than the mean in the patches. For each pixel $(x,y)$, its sparse selection probability follows the Bernoulli distribution. The process can be described as following formula:
\begin{equation}
S(x, y)= 
\begin{cases}
F_{\mathrm{i}}^{\mathrm{R}},&\text{if}~M_{k}(x,y)>\frac{M_{k}(x, y)}{\sum_{(x, y) \in N_{h, w}^{p}\left|N_{x, y}^{p}\right|}}\\
0,&\text{otherwise}	
\end{cases}
\end{equation}
where $S(x,y)$ is the selected sparse pixel, and  where $M_{k}(x,y)$ is the pixel value in the uncertainty score. The sparse optimization strategy can select high-frequency features from sparse industrial defect features so as to more accurately describe the contour and other details of the object. However, the sparse selection mechanism lacks effective supervision, resulting in a small number of selected feature points deviating from the object. Therefore, we introduce an IoU-based optimization mechanism to optimize the feature activation mechanism iteratively. The goal of the optimization mechanism is to make the selected high-frequency pixels fall within the ground truth area as much as possible. In practice, the optimization mechanism uses an additional detection head for regression prediction and optimizes the effect of feature activation through CIoU.
\subsection{Text-Visual Mutual} 
\label{Text-Visual Mutual}
\textbf{Motivation}: Industrial defects only constitute a small portion of the image. Directly performing a simple dot product between visual features and text embeddings would introduce significant background noise. Therefore, we use text-guided multi-scale maximal sigmoid attention to focus text keywords onto the spatial locations corresponding to defects, generating a consistent text-visual feature map that suppresses responses from irrelevant regions.

As shown in Fig.~\ref{Architect} (c), given the text embedding $T_{i}$ from text encoder and the image feature $\mathrm{F}_{\mathrm{L}} \in \mathrm{R}^{\mathrm{C\times H\times W}}(\mathrm{L} \in\{1,2,3,4\})$. We adopt multi-scale image features and aggregates the text features into the image features using the maximum Sigmoid attention query text-image matching semantic features. The definition of $\mathrm{F}_{\rm{img-text}}$ reads as:
\begin{equation}
\label{eq_F}
\mathrm{F}_{\rm{img-text }}=\mathrm{F}_{\mathrm{L}} \times Sigmoid\left(\max\left(\mathrm{F}_{\mathrm{L}} \times \mathrm{T}_{\mathrm{i}}^{\mathrm{\top}}\right)\right)^{\mathrm{\top}}.
\end{equation}
Note that the notation $\max(\mathrm{F}_{\mathrm{L}} \times \mathrm{T}_{\mathrm{i}}^{\mathrm{\top}})$ in Eq.~\eqref{eq_F} means to find the maximum value of each column of the matrix $\mathrm{F}_{\mathrm{L}} \times \mathrm{T}_{\mathrm{i}}^{\mathrm{\top}}$.

\subsection{Visual-Text Mutual}
\label{Visual-Text Mutual}
\textbf{Motivation}: Text-Visual Mutual only implements a one-way weighting of text onto visual features, failing to compensate for missing spatial details in the text. Therefore, we calculate the inverse pixel-level similarity between image and text and achieve image-to-text semantic alignment. This enhances the expression of defect location and semantics for simple category (such as “crack”), alleviating overly coarse semantics in open-category text and misalignment with images.

As shown in the Fig.~\ref{Architect} (d), the text embedding is padded and adjusted to the same dimension as the image feature. The purpose is to introduce the same spatial dimension as image features into the text embedding of simple words, making it easier to introduce more fine-grained visual information. Then, the spatial and channel dimensions of the image feature are scored with the text feature for similarity, and the scoring can be described as follows:
\begin{equation}
\begin{array}{c}
\text {Score}=\sum_{1}^{C} \sum_{1}^{H} \sum_{1}^{W} \frac{\sum_{i=1}^{n} I_{i} T_{i}}{\sqrt{\sum_{i=1}^{n} I_{i}^{2}} \times \sqrt{\sum_{i=1}^{n} T_{i}^{2}}},
\end{array}
\end{equation}
where $I_{i}$ is the image feature and $T_{i}$ is the text feature. Finally, the image feature is multiplied by the similarity score and added to the text to obtain the feature with image-text semantics. Different from the traditional cross-attention-based method \cite{yoloworld,liu2023groundingdino}, we use similarity scores as weights to achieve pixel-level semantic alignment of image features to text features. Thus, it forms image-guided text enhancement and compensates for the missing visual details (such as spatial relationships of objects) in simple text descriptions.
\subsection{Downstream task applications} 
Based on the RTVPNet, we define three downstream tasks for industrial scenarios, as illustrated in Fig.~\ref{Architect}. For the Visual Grounding task, the refined visual-text features are applied for multi-scale feature regression positioning, and the loss function are constructed by CIoU \cite{zheng2020distance}. For the defect detection task, we introduce Info-NCE \cite{InfoNce} to optimize the visual-text feature matching. For the Vision-Question-Answer task, we introduce DeepSeek \cite{2025DeepSeek} for Visual Question Answering. Among them, the DeepSeek parameters are completely frozen because the detection model pre-trained for industrial scenarios can give DeepSeek more professional industrial scenario knowledge, and there is no need to retrain DeepSeek.

\section{Experiments}
\label{sec_Exp}

First of all we extensively evaluate the proposed MMIOC-1M to demonstrate its effectiveness, and then test the performance of the RTVPNet on MMIOC-1M closed and open related tasks.

\subsection{Dataset and Evaluation Metrics}
We conduct the experiments on the MMIOC-1M, MSCOCO \cite{COCO} and LVIS \cite{LVIS}. For the MMIOC-1M open scenes, we conduct experiments on the part of the MMIOC-1M, which contains a total of 94 visible classes and 64 invisible classes. For the MMIOC-1M closed scene task, we split all 1M samples of MMIOC-1M into an 80\% training set and a 20\% test set. To evaluate the generalization ability, we perform closed scene verification on the COCO dataset and open-vocabulary verification on the LVIS dataset. We use COCO and LVIS indicators to measure the model's accuracy. For the LVLMs comparison, we added an additional detection module to fine-tune its output tokens. Furthermore, we use Parameters and GFLOPs to measure the every modules total effective parameters and computational cost of RTVPNet.

\subsection{Implementation Details}
The model is built on PyTorch 2.0.1, and the hardware environment is 8 Nvidia A100 GPUs. The model is trained 200 epochs using AdamW with 128 batches. The input image size is 640. The initial learning rate is 2e-3, the weight decay is 0.025, the text encoder (CLIP-Text Encoder) and Mobile-SAM-T's encoder are frozen during pre-training. The expert model is trained on MMIOC-1M visible classes for open task. We use different CNN models as the expert model of RTVPNet. Unless otherwise specified, RTVPNet-S uses C2F-S \cite{YOLOv8} by default, and RTVPNet-L uses C2F-L \cite{YOLOv8} by default.

\subsection{Quantitative experiments with the State-of-the-art}
\subsubsection{MMIOC-1M quality analysis} 

We verify the accuracy various detection models to demonstrate MMIOC-1M quality and practicality. We focus on the analysis of recently proposed open-vocabulary and one-stage models that are widely approved for their ability to handle open and closed scenes. As shown in Table \ref{MMIOCOPen}, the AP of the recently proposed models for open scenes is more than 11\%, which proves the effectiveness of the MMIOC-1M open scene data. However, the AP of all models is less than 15\%, indicating that MMIOC-1M is a challenging open benchmark. As shown in Table \ref{MMIOCClosed}, the average AP of the general and defect detection models exceeds 15\%. More advanced models further improve the effect. In addition, the general object detection model performs poorly. This is because MMIOC-1M has 351 categories of industrial defects, and there are both inter-class similarities and intra-class differences. It is difficult for general object detection models to capture complex scene features, which also illustrates the necessity of multi-modal interaction. In summary, the performance of various models on MMIOC-1M indirectly proves the effectiveness of MMIOC-1M annotation and image quality.

\subsubsection{MMIOC-1M open scenario} 
As shown in Table \ref{MMIOCOPen}, the open-scene experiments of MMIOC-1M demonstrate that RTVPNet achieves leading detection accuracy with low computational and parameter consumption. Compared to general open detectors, RTVPNet-S improves AP50 to 26.7\% with 76 M parameters and 17 GFLOPs, 2.7\% higher than YOLO-World-S \cite{yoloworld} of the same scale, and 3.4\% higher than the latest DOSOD-S \cite{DOSOD}. RTVPNet-L further improves AP50 to 30.7\%, still consuming only 110 M parameters and 89 GFLOPs, significantly lower than YOLO-UNIOW-L \cite{liu2024yoloUNI} and Grounding CLIPv2-T \cite{liu2023groundingdino}. Compared to LVLMs, which require 3-8 B parameters to training, RTVPNet-S achieves equal or even better performance with approximately 1/40th the number of parameters. Experimental results validate that RTVPNet’s proposed Refined Visual Prompt and cross-modal bidirectional interaction mechanism for industrial open scenarios effectively improve generalization to unseen categories while significantly reducing computational overhead. Notably, all compared methods achieved an mAP of less than 20\%, highlighting the challenges of MMIOC-1M and facilitating future industrial open scenario tasks.

\begin{table}[t]
\caption{Comparative experiment of MMIOC-1M in open scene.}
\fontsize{16}{18}\selectfont\rmfamily
\centering
\resizebox{1\columnwidth}{!}{
\begin{tabular}{ccccccccc}
\hline
Method                   & AP  & AP50 & AP75 & APs & APm  & APl  & Param  & GFLOPs     \\\hline
\multicolumn{9}{c}{\textbf{Open Detection Based}}                                           \\\hline
Mamba-YOLO-World-S \cite{Mamba-YOLO-World}      & 12.7 & 24.3 & 10.5 & 4.4 & 8.3  & 13.4 & 78M         & 297        \\
DOSOD-S \cite{DOSOD}                 & 11.4 & 22.6 & 9.7  & 5.9 & 10.6 & 11.8 & 76M         & 15         \\
Grounding Dino-T \cite{liu2023groundingdino}        & 12.0 & 21.9 & 8.1  & 6.4 & 11.2 & 12.7 & 172M        & -          \\
YOLO-UNIOW-S \cite{liu2024yoloUNI}            & 14.2 & 25.8 & 11.6 & 6.3 & 12.5 & 11.7 & 73M         & 12         \\
YOLO-World-S \cite{yoloworld}            & 12.2 & 23.3 & 12.6 & 7.9 & 12.7 & 14.0 & 77M         & 297        \\
Grounding CLIPv2-T      & 12.0 & 22.8 & 11.9 & 7.3 & 12.1 & 13.7 & 232M        & -          \\
Mamba-YOLO-World-M       & 14.8 & 27.1 & 12.8 & 6.4 & 11.8 & 15.1 & 94M         & 324        \\
Mamba-YOLO-World-L       & 14.6 & 26.8 & 12.8 & 6.9 & 11.0 & 14.8 & 113M        & 369        \\
YOLO-UNIOW-M             & 15.0 & 28.2 & 13.9 & 6.1 & 13.2 & 14.0 & 82M       & 32         \\
YOLO-UNIOW-L             & 16.6 & 30.3 & 15.1 & 7.5 & 14.6 & 15.6 & 95M         & 70         \\
YOLO-World-M             & 14.5 & 27.2 & 13.6 & 7.2 & 13.9 & 15.6 & 92M         & 324        \\
YOLO-World-L             & 16.1 & 28.4 & 14.6 & 8.5 & 14.7 & 16.0 & 111M        & 370        \\ 
\rowcolor[HTML]{EFEFEF} 
RTVP (Previous Work) \cite{Zero-shotLearning} & 14.0 & 25.1 & 12.4 & 8.1 & 13.2 & 14.0 & 131M        & 39         \\\hline
\multicolumn{9}{c}{\textbf{LVLMs Based}}                                                    \\\hline
Qwen2.5-VL 3B \cite{2025Qwen2}          & 15.9 & 28.1 & 14.5 & 8.3 & 13.6 & 15.3 & \textbf{-} & \textbf{-} \\
Qwen3-VL 4B \cite{2025Qwen3}            & 14.6 & 27.2 & 12.7 & 8.1 & 12.5 & 13.2 & \textbf{-} & \textbf{-} \\
DefectGLM 7.2B \cite{defectglm}         & 13.8 & 24.7 & 13.2 & 6.4 & 12.0 & 11.6 & -          & -          \\
LLaVA-NeXT 7B \cite{Llava-next}           & 11.7 & 22.0 & 10.2 & 6.6 & 10.9 & 10.4 & \textbf{-} & \textbf{-} \\
LLaVA-OV 7B \cite{2024LLaVA-ov}             & 13.9 & 26.9 & 12.3 & 9.7 & 14.1 & 13.0 & \textbf{-} & \textbf{-} \\\hline
\rowcolor[HTML]{E6E0E0} 
RTVPNet-S                & 14.9 & 26.7 & 13.6 & 8.3 & 13.6 & 14.3 & 76M         & 17         \\
\rowcolor[HTML]{E6E0E0} 
RTVPNet-L                & 17.4 & 30.7 & 15.9 & 9.7 & 15.9 & 16.8 & 110M        & 89        \\\hline
\end{tabular}}
\label{MMIOCOPen}
\end{table} 

\subsubsection{LVIS Open Scene Generalization} 
Table \ref{Lviscomp} evaluates the generalization ability of RTVPNet in open-vocabulary transfer on LVIS \cite{LVIS}. Specifically, RTVPNet uses the YOLOv11 \cite{yolov11} as the expert model, is pre-trained on the Object365 and GoldG. Compared with the latest YOLO-World, RTVPNet improves by 3.2\% in AP. YOLO-World \cite{yoloworld}, GLIP \cite{glip}, and GroundingDino \cite{liu2023groundingdino} are typical open-vocabulary detectors based on single-text prompts. The gain of RTVPNet proves that the refined visual prompts and the interaction of visual-text prompts effectively improve expert's ability to understand open scenes. It is worth noting that the performance of RTVPNet has been further improved compared to the our earlier version \cite{Zero-shotLearning}. Compared with other open detectors, the AP of RTVPNet is still improved, proving that RTVPNet can be generalized to general open detection scenarios. In addition, with the increase of pre-training data, the performance of RTVPNet has been improved, indicating that pre-training with a large amount of data has a positive effect on accurate prompt expression.

\begin{table}
\caption{LVIS open scene generalization, where different datasets  are exploited for pre-training.}
\fontsize{16}{18}\selectfont\rmfamily
\centering
\resizebox{1\columnwidth}{!}
{
\begin{tabular}{ccccccc}
\hline
Method             & Data       & Param & AP   & APr  & APc  & APf  \\\hline
GLIP-T \cite{glip}            & O365       & 232M   & 17.8 & 13.5 & 12.8 & 22.2 \\
YOLO-World-v1 \cite{yoloworld}        & O365       & 77M    & 23.5 & 16.2 & 21.1 & 27.0 \\
ViLD \cite{gu2021open}              & O365       & -     & -    & -    & 20.0 & 28.3 \\
                                                \rowcolor[HTML]{EFEFEF}
RTVP (Previous work)           & O365       & 131M   & 24.1 & 17.3 & 22.4 & 27.9 \\
\rowcolor[HTML]{E6E0E0} 
RTVPNet (YOLOv11s) \cite{Zero-shotLearning}           & O365       & 73   & 25.2 & 18.3 & 23.2 & 28.5 
\\\hline
Grounding-Dino-T \cite{liu2023groundingdino}  & O365,GlodG & 173   & 25.6 & 14.4 & 19.6 & 32.2 \\
GLIP-T             & O365,GlodG & 232   & 24.9 & 17.7 & 19.5 & 31.0 \\
Mamba-YOLO-World-S & O365,GlodG & 78    & 27.7 & 19.5 & 27.0 & 29.9 \\
YOLO-World-S       & O365,GlodG & 77    & 26.2 & 19.1 & 23.6 & 29.8 \\
\rowcolor[HTML]{EFEFEF}
RTVP (Our Previous work) \cite{Zero-shotLearning}  & O365,GlodG  & 131   & 26.8 & 19.5 & 23.4 & 30.7 \\
\rowcolor[HTML]{E6E0E0} 
RTVPNet (YOLOv11s) \cite{Zero-shotLearning}      & O365,GlodG & 73   & 29.4 & 22.8 & 27.8 & 31.9 \\\hline
\end{tabular}}
\label{Lviscomp}
\end{table}

\subsubsection{MMIOC-1M Closed Scenario} 
Table \ref{MMIOCClosed} compares the general model and the expert defect model in MMIOC-1M closed scenarios. The closed scene data of MMIOC-1M exhibits a long-tail distribution, yet RTVPNet still achieves SOTA. RTVPNet outperforms the traditional expert model by nearly 15\%, indicating that RTVPNet is highly sensitive to large-scale long-tail distribution datasets. Compared to the user training-detection mode of traditional expert models, RTVPNet supports user-defined vocabulary and automatically generates refined prompts, providing high accuracy. Compared with the YOLO-PCB \cite{YOLO-PCB} defect-specific model and the recently proposed YOLOv12L \cite{tian2025yolov12}, RTVPNet achieves a 15\% improvement in AP while maintaining a GFLOPs advantage. RTVPNet also achieves optimal performance compared to LVLMs such as Qwen3-VL 4B. Experiments show that RTVPNet can also improve accuracy in closed scenarios. Because RTVPNet introduces an expert model and refines visual prompts and text interactions, it can transfer richer knowledge to the VLM.

\begin{table}[t]
\caption{Comparative experiment of MMIOC-1M in closed scene.}
\fontsize{16}{18}\selectfont\rmfamily
\centering
\resizebox{1\columnwidth}{!}{
\begin{tabular}{ccccccccc}
\hline
Method                   & AP   & AP50 & AP75 & APs  & APm  & APl  & Param & GFLOPs \\ \hline
\multicolumn{9}{c}{\textbf{Closed Detection Based}}                                      \\\hline
YOLOv12S \cite{tian2025yolov12}                & 23.5 & 47.8 & 22.7 & 13.6 & 21.4 & 32.3 & 9M        & 21     \\
YOLOv10S  \cite{wang2024yolov10}               & 20.8 & 39.4 & 18.6 & 9.0  & 15.9 & 22.8 & 7M        & 22     \\
YOLOv8S \cite{YOLOv8}                 & 9.9  & 19.5 & 8.4  & 4.0  & 7.1  & 10.0 & 11M       & 29     \\
Mamba-YOLO-B \cite{wang2024mambayolo}            & 10.1 & 22.5 & 9.9  & 4.3  & 9.0  & 12.4 & 19M       & 45     \\
Hyper-YOLO-T \cite{feng2024hyper}            & 13.4 & 26.0 & 11.4 & 5.3  & 9.7  & 13.6 & 3M        & 10     \\
Lite-YOLO-ID \cite{lite-yolo}            & 14.1 & 26.3 & 11.2 & 6.1  & 10.6 & 15.9 & 4M        & 9      \\
YOLO-PCB \cite{YOLO-PCB}                & 12.5 & 24.8 & 10.6 & 4.8  & 9.5  & 13.3 & 15M       & 20     \\
LF-YOLO-1.25 \cite{lf}             & 23.5 & 41.2 & 21.6 & 11.1 & 18.7 & 25.9 & 8M        & 25      \\
ETDNet \cite{Etdnet}                  & 17.1 & 34.2 & 14.4 & 14.0 & 24.2 & 32.2 & 7M        & 24     \\
SSA-YOLO \cite{huang2024ssa}                & 21.3 & 36.9 & 15.1 & 14.7 & 19.5 & 36.4 & 13M       & 18     \\
YOLOv12L                 & 37.8 & 58.4 & 31.4 & 16.3 & 27.2 & 42.7 & 26M       & 89     \\
YOLOv12M                 & 31.1 & 54.3 & 27.5 & 15.9 & 25.8 & 39.1 & 20M       & 68     \\
YOLOv10M                 & 21.3 & 41.2 & 19.0 & 10.6 & 17.9 & 22.7 & 15M       & 59     \\
YOLOv10L                 & 23.9 & 44.8 & 22.3 & 11.7 & 20.6 & 26.1 & 24M       & 120    \\
YOLOv8M                  & 20.8 & 37.6 & 19.3 & 9.5  & 16.6 & 21.9 & 26M       & 79     \\
YOLOv8L                  & 21.9 & 38.1 & 19.8 & 10.5 & 17.0 & 22.8 & 44M       & 165    \\
MDETR \cite{kamath2021mdetr}                   & 22.6 & 39.0 & 21.7 & 11.0 & 19.8 & 25.6 & 169M      & -      \\ 
\rowcolor[HTML]{EFEFEF}
RTVP (Previous Work) & 33.0 & 55.9 & 33.4 & 14.5 & 26.6 & 36.8 & 131M      & 39     \\\hline
\multicolumn{9}{c}{\textbf{LVLMs Based}}                                             \\ \hline
DefectGLM 7.2B \cite{defectglm}          & 27.1 & 51.5 & 26.6 & 13.3 & 26.3 & 37.9 & -        & -      \\
Qwen3-VL 4B \cite{2025Qwen3}             & 38.9 & 61.8 & 37.3 & 16.7 & 31.8 & 43.6 & -        & -      \\
LLaVA-OV 7B \cite{2024LLaVA-ov}             & 36.6 & 60.2 & 35.5 & 14.0 & 30.4 & 42.4 & -        & -      \\\hline
\rowcolor[HTML]{E6E0E0} 
RTVPNet-S                & 36.7 & 60.7 & 36.5 & 16.7 & 30.4 & 40.9 & 76M       & 17     \\
\rowcolor[HTML]{E6E0E0} 
RTVPNet-L                & 41.3 & 64.9 & 38.5 & 19.2 & 34.1 & 45.9 & 110M      & 89    \\\hline
\end{tabular}}
\label{MMIOCClosed}
\end{table}

\subsubsection{COCO Closed Scene Generalization} 

Table \ref{COCOcomp} verifies the ability of RTVPNet to generalize in closed scenes on MS-COCO \cite{COCO}. When the comparison method is a more advanced network, the detection performance consistently improves. We introduce YOLOv11 \cite{yolov11} as a baseline to verify the effectiveness of RTVPNet in improving the performance of traditional expert models. Specifically, we replace the expert model of RTVPNet with YOLOv11's S and L versions of backbone. We compare YOLO-World \cite{yoloworld} and GroundingDino's \cite{liu2023groundingdino} pre-trained versions on Object365 \cite{shao2019objects365}, CC3M and GoldG. Compared with YOLOv8-v10, REVPNet's AP is improved by 3\% on average. The gain is due to RTVPNet's introduction of Mobile-SAM as a knowledge prior to improving the model feature extraction ability. Compared with the baseline YOLOv11, RTVPNet significantly improves multi-scale AP (about 1\%). The reason is that the refined visual prompts enhance multi-scale objects, and the text interaction further refines the semantic information of multi-scale objects. In summary, the results further illustrates that RTVPNet can generalize and improve the accuracy of traditional expert models in closed scenarios. Compared with YOLO-World \cite{yoloworld} and GroundingDino \cite{liu2023groundingdino}, which are pre-trained in open scenes, all indicators of RTVPNet have reached the best. Different from general open detectors, RTVPNet has a unique expert-guided domain projection, refined visual prompt, and visual-text bidirectional interaction, giving RTVPNet a unique advantage.
\begin{table}
\caption{COCO closed scene generalization.}
\fontsize{4}{5}\selectfont\rmfamily
\centering
\resizebox{0.9\columnwidth}{!}{
\begin{tabular}{cccc}
\hline
Method    & AP   & AP50 & AP75 \\\hline
YOLOv7-T \cite{wang2022yolov7}          & 37.5          & 55.8          & 40.2          \\
YOLOv7-L           & 50.9          & 69.3          & 55.3          \\\hline
YOLOv8-S  \cite{YOLOv8}         & 44.4          & 61.2          & 48.1          \\
YOLOv8-L           & 52.9          & 69.9          & 57.7          \\\hline
\textbf{YOLOv11-S} \cite{yolov11} & \textbf{46.9} & \textbf{63.9} & \textbf{50.6} \\
\textbf{YOLOv11-L} & \textbf{53.3} & \textbf{70.1} & \textbf{58.2} \\\hline
GroundingDino-T \cite{liu2023groundingdino}   & 48.4          & -             & -             \\
GroundingDino-L    & 60.7          & -             & -             \\\hline
YOLO-World-S \cite{yoloworld}      & 45.7          & 62.3          & 49.9          \\
YOLO-World-L       & 53.3          & 70.3          & 58.1          \\\hline
\rowcolor[HTML]{E6E0E0} 
RTVPNet (YOLOv11s) & 48.2 & 64.1 & 51.7 \\\hline
\rowcolor[HTML]{E6E0E0} 
RTVPNet (YOLOv11l)& 54.2 & 70.4 & 58.3\\\hline
\end{tabular}}
\label{COCOcomp}
\end{table}\\

\begin{figure*}[!h]
\centering
\includegraphics[width=6in]{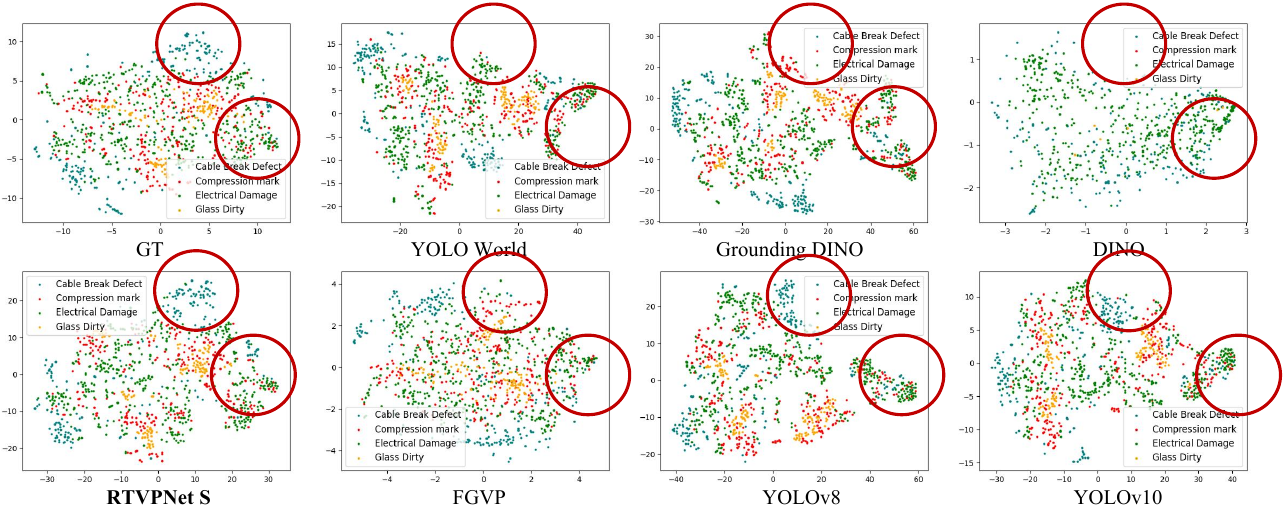}
\caption{T-SNE visualization in MMIOC-1M open scenes. Compared with Ground Truth and other methods, RTVPNet can generate the most compact feature representation, which effectively promotes learning about invisible categories in open scenes.}
\label{tsne}
\end{figure*}

\subsection{Qualitative Experiments with State-of-the-art}
To further demonstrate the effectiveness of RTVPNet in optimizing the distribution of visual features, we use t-SNE \cite{t-sne} to visualize the invisible features of RTVPNet-S in the open scene of MMIOC-1M. FGVP \cite{yang2024fine} is reproduced with YOLO-World \cite{yoloworld} as the baseline. As shown in the Figure \ref{tsne}, compared with using Ground Truth and FGVP \cite{yang2024fine} (visual prompts only), the features of RTVPNet show clear clustering, demonstrating the importance of fine-grained visual and textual prompts interaction. Compared with other open detectors guided by text features, RTVPNet has more consistent clustering with Ground Truth, indicating that RTVPNet produces features that are more consistent with actual classification, generates well-separated clusters for different classes. Compared with other closed scene detectors, although they produce tighter clusters, they deviate from the factual distribution of Ground Truth. It is worth noting that the categories Cable Break and Electronic Damage selected from MMIOC-1M have inter-class similarities, but RTVPNet can still accurately generate different clusters for similar categories. In summary, t-SNE visualization shows that RTVPNet can optimize the distribution of invisible category features in open scenarios, effectively promoting the learning of invisible categories in open scenarios.

\begin{figure*}[!h]
\centering
\includegraphics[width=6in]{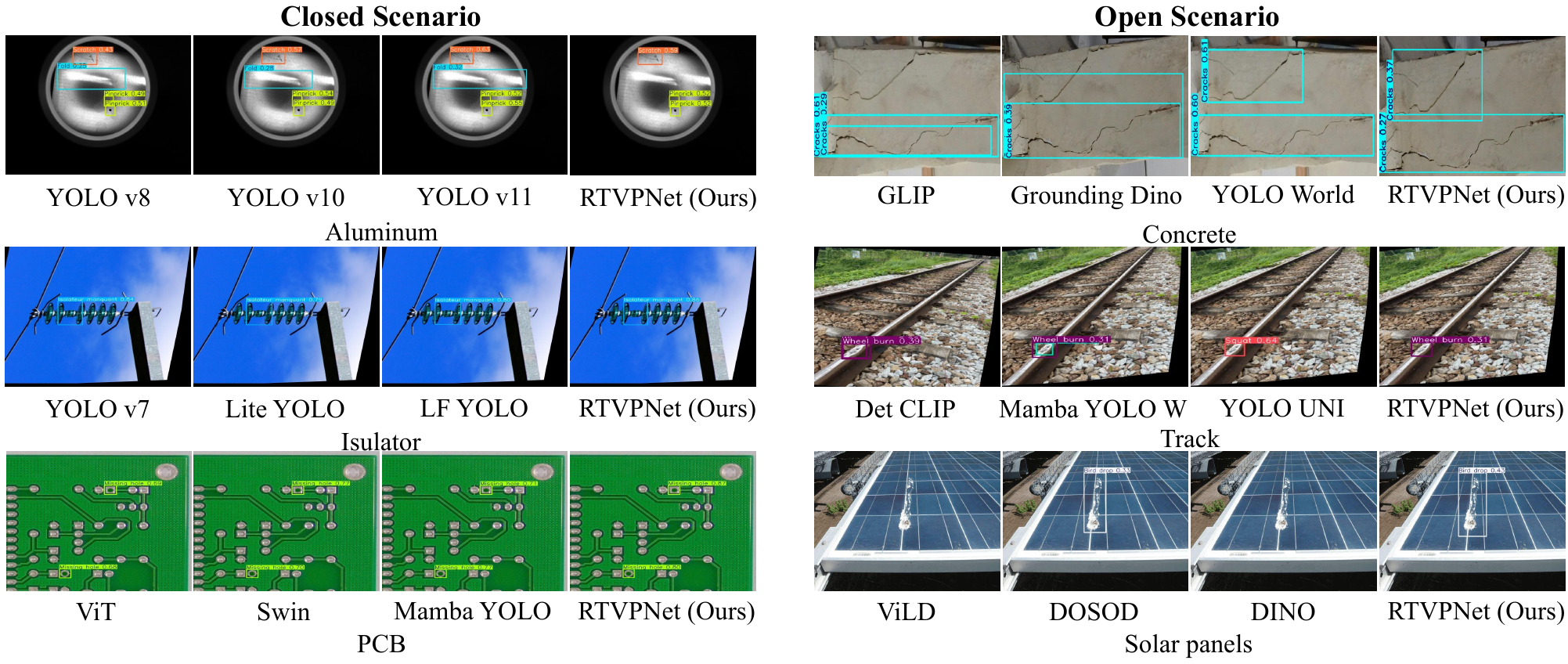}
\caption{Comparison of detection results in open and closed scenarios of MMIOC-1M. RTVPNet has significant advantages in closed and open scenarios. Additionally, the results also demonstrate that MMIOC-1M is a challenging benchmark.}
\label{Det}
\end{figure*}

To further evaluate the detection effectiveness of RTVPNet on MMIOC-1M, we visualize the results of RTVPNet-S on MMIOC-1M open and closed scenes. For all detection result comparisons, the left column is the results of closed scenes, and the right column is the results of open scenes. As shown in the closed scene comparison in Fig.~\ref{Det}, the existing defect detection methods (Lite YOLO \cite{lite-yolo}, LF YOLO \cite{lf}) have lower object confidence for isolators, while RTVPNet provides higher object confidence. Compared with existing general closed scene detectors, YOLOv8 \cite{YOLOv8} and YOLOv11 \cite{yolov11} mistakenly identify reflective areas in aluminum as defects. Different from the above methods, RTVPNet designs three unique innovations, which has higher classification confidence and good robustness in complex noisy scenes.

As shown in the open scene comparison in Fig.~\ref{Det}, due to the large differences in the scenes and representations of visible and invisible categories, it is very challenging to identify invisible categories. The introduction of large pre-trained model in RTVPNet can enable VLM to obtain better feature classification and positioning capabilities. For example, in the first row, RTVPNet shows a more accurate defect feature positioning capability. The second and third rows illustrate that the baseline method encounters issues of mis-classification (railway defects in the second row) and low recall (defects not detected in the third row). It is worth noting that RTVPNet can still accurately detect defects with drastic scale changes. In summary, RTVPNet maintains the accuracy of positioning and classification when detecting defects.

\begin{figure}[!h]
\centering
\includegraphics[width=3.3in]{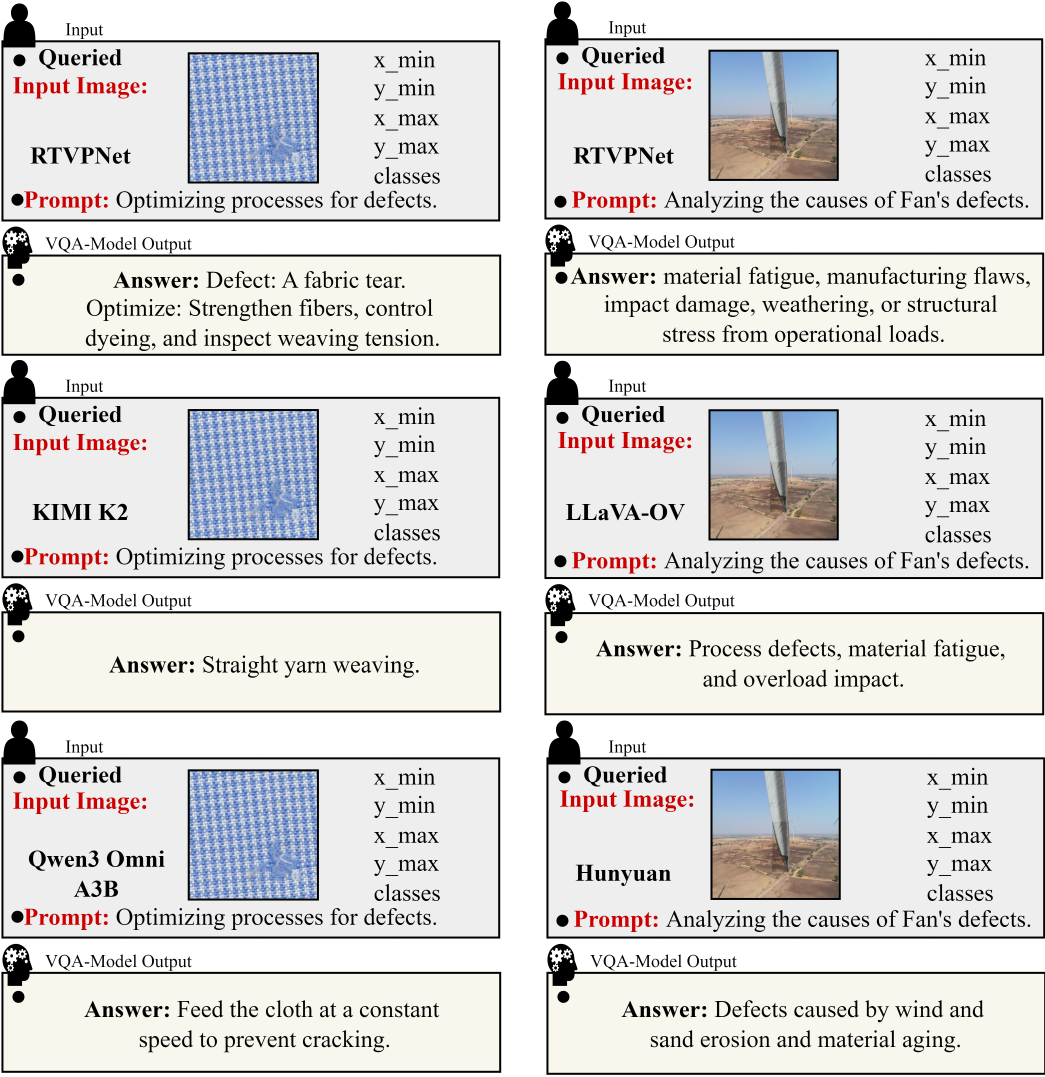}
\caption{Visual Question Answering results on MMIOC-1M. The combination of RTVPNet and Deepseek performs exceptionally well in qualitative analysis and descriptive capabilities. It also illustrates the accuracy of RTVPNet combined with MMIOC-1M in industrial feature extraction.}
\label{VQA}
\end{figure}

As shown in Fig.~\ref{VQA}, we evaluated the performance of RTVPNet and other LVLMs on the VQA task. Traditional LVLMs have limited insights, while RTVPNet integrates the Deepseek model to demonstrate powerful qualitative analysis and description capabilities. The core advantage of RTVPNet lies in its ability to detect defects and analyze features in detail accurately. Taking the "fabric tear" image as an example, the model deeply analyzes the characteristics and causes of defects, subverting the traditional isolated prediction model. RTVPNet breaks the boundaries between image and text analysis through multi-modal interaction, providing engineers with instant and detailed information. Overall, MMIOC-1M and RTVPNet have become a groundbreaking innovation in industrial detection, characterized by interactivity, high information volume, and comprehensiveness, which accurately detect defects, enhances feedback, and improves work efficiency and accuracy.

\subsection{Ablation Study}

\begin{table}
\caption{Ablation experiments of components on MMIOC-1M. EDP is Expert-assisted Domain Projection, RVP is Refined Visual Prompt, V-T is Visual-Text mutual, and T-V is Text-Visual mutual}
\fontsize{4}{5}\selectfont\rmfamily
\centering
\resizebox{1\columnwidth}{!}{
\begin{tabular}{ccccccc}
\hline
\multicolumn{7}{c}{RTVPNet-L (Open)}                                                                                     \\\hline
EDP & RVP & V-T & T-V& AP   & AP50 & AP75 \\\hline
$\surd$                           &                       &                    &                    & 14.7 & 27.5 & 11.8 \\
                                  & $\surd$               &                    &                    & 12.8 & 24.9 & 10.1 \\
                                  &                       & $\surd$            &                    & 13.3 & 25.2 & 9.9  \\
                                  &                       &                    & $\surd$            & 13.7 & 25.7 & 10.2 \\\hline
$\surd$                           & $\surd$               &                    &                    & 15.3 & 28.1 & 12.2 \\
$\surd$                           &                       & $\surd$            &                    & 16.0 & 29.0 & 14.1 \\
$\surd$                           &                       &                    & $\surd$            & 16.2 & 28.7 & 14.5 \\
                                  & $\surd$               & $\surd$            &                    & 15.2 & 27.9 & 13.3 \\
                                  & $\surd$               &                    & $\surd$            & 15.0 & 27.6 & 12.9 \\
                                  &                       & $\surd$            & $\surd$            & 14.0 & 26.4 & 10.5 \\\hline
$\surd$                           & $\surd$               & $\surd$            &                    & 15.6 & 28.7 & 12.6 \\
                                  & $\surd$               & $\surd$            & $\surd$            & 15.7 & 28.6 & 13.7 \\
$\surd$                           &                       & $\surd$            & $\surd$            & 16.5 & 29.7 & 14.6 \\
$\surd$                           & $\surd$               &                    & $\surd$            & 15.9 & 29.2 & 13.0 \\\hline
$\surd$                           & $\surd$               & $\surd$            & $\surd$            & 17.4 & 30.7 & 15.9 \\\hline
\multicolumn{7}{c}{RTVPNet-S (Closed)}                                                                                   \\\hline
$\surd$                           &                       &                    &                    & 32.6 & 55.2 & 31.4 \\
                                  & $\surd$               &                    &                    & 30.5 & 53.1 & 29.3 \\
                                  &                       & $\surd$            &                    & 31.3 & 53.7 & 29.9 \\
                                  &                       &                    & $\surd$            & 31.6 & 53.2 & 30.0 \\\hline
$\surd$                           & $\surd$               &                    &                    & 33.1 & 56.4 & 31.7 \\
$\surd$                           &                       & $\surd$            &                    & 32.9 & 55.8 & 32.1 \\
$\surd$                           &                       &                    & $\surd$            & 32.8 & 56.0 & 32.1 \\
                                  & $\surd$               & $\surd$            &                    & 30.8 & 53.7 & 29.4 \\
                                  & $\surd$               &                    & $\surd$            & 31.1 & 53.8 & 29.4 \\\hline
                                  &                       & $\surd$            & $\surd$            & 31.8 & 54.4 & 30.6 \\
$\surd$                           & $\surd$               & $\surd$            &                    & 33.4 & 56.8 & 32.0 \\
                                  & $\surd$               & $\surd$            & $\surd$            & 34.0 & 56.9 & 32.5 \\
$\surd$                           &                       & $\surd$            & $\surd$            & 34.5 & 57.9 & 33.8 \\
$\surd$                           & $\surd$               &                    & $\surd$            & 33.6 & 57.0 & 32.1 \\\hline
$\surd$                           & $\surd$               & $\surd$            & $\surd$            & 36.7 & 60.7 & 36.5 \\\hline
\end{tabular}}
\label{Compnet}
\end{table}
\subsubsection{Ablation Experiments of Components} 

Table \ref{Compnet} shows the results of ablation studies of different components on MMIOC-1M. Expert-assisted Domain Projection can effectively improve the accuracy of closed and open scenes. Since the expert model can provide expert knowledge supervision to Mobile-SAM, the migration effect in the industrial field is enhanced. Refined Visual Prompts are more conducive to industrial detection in open scenes. Because it uses sparse modelling of industrial images to help focus on key features. Cross-modal text-visual interaction helps to further refine semantic features related to objects, which is conducive to accurately extracting invisible class features.

\subsubsection{Expert-guided Domain Projection} 
As described in section \ref{Expert-assisted Domain Projection} regarding motivation, we provide domain expert knowledge for Mobile-SAM to improve its generalization ability in industrial scenarios. Expert-guided Domain Projection includes pre-training of expert models, domain-aware projection, and dynamic updating of expert models. As shown in Table \ref{expertdomain}, it can be seen from the experiments of MMIOC-1M open scenarios that pre-training of expert models is crucial under class-invisible conditions. Pre-training can effectively inject more industrial expert knowledge into Mobile-SAM. Adding domain-aware projection and dynamic expert updating further improves the detection effect of invisible categories. This is because the iterative optimization of the expert model can improve the adaptability of the expert model to industry scenarios, thereby better guiding the industry scenario domain migration of SAM. The experiments on the closed scenes of MMIOC-1M show that pre-training and updating the expert model in the visible category is also a crucial step. Domain projection further improves the accuracy of closed scene domain adaptation. In summary, the expert-guided domain projection components can improve the generalization ability and robustness of Mobile-SAM in industrial scenarios. Therefore, expert-guided domain projection can be used for SAM's industrial scene task migration.

\begin{table}
\caption{Expert-guided Domain Projection Ablation Experiments on MMIOC-1M.}
\centering
\fontsize{5}{6}\selectfont\rmfamily
\resizebox{0.9\columnwidth}{!}{
\begin{tabular}{ccccc}
\hline
\multicolumn{5}{c}{MMIOC-1M Open Scene (RTVPNet-S)}                              \\\hline
Pretrain             & Projection           & Update               & AP   & AP50 \\\hline
$\surd$ &                      &                      & 11.9 & 24.0 \\
                     & $\surd$ &                      & 11.5 & 22.3 \\
                     &                      & $\surd$ & 8.2  & 11.1 \\\hline
$\surd$ & $\surd$ &                      & 14.7 & 25.8 \\
$\surd$ &                      & $\surd$ & 14.4 & 26.0 \\
                     & $\surd$ & $\surd$ & 14.0 & 25.6 \\\hline
$\surd$ & $\surd$ & $\surd$ & 14.9 & 26.7 \\\hline
\multicolumn{5}{c}{MMIOC-1M Closed Scene (RTVPNet-L)}                            \\\hline
Pretrain             & Projection           & Update               & AP   & AP50 \\\hline
$\surd$ &                      &                      & 36.8 & 58.4 \\
                     & $\surd$ &                      & 37.5 & 59.6 \\
                     &                      & $\surd$ & 32.6 & 54.2 \\\hline
$\surd$ & $\surd$ &                      & 39.3 & 61.9 \\
$\surd$ &                      & $\surd$ & 38.4 & 60.8 \\
                     & $\surd$ & $\surd$ & 38.7 & 61.1 \\\hline
$\surd$ & $\surd$ & $\surd$ & 41.3 & 64.9 \\\hline
\end{tabular}}
\label{expertdomain}

\end{table}

Domain-aware projection uses feature masking and reconstruction to get more accurate industrial feature expressions. We verify the impact of different mask ratio. Table \ref{Domainmask} verifies the effect of mask ratio in MMIOC-1M. Experiments show that the detection effect is best when the mask ratio is 20\%. Too small mask ratio will cause Mobile-SAM to fail to learn key feature representations. Too large mask ratio will cause distortion or loss of key features.

\begin{table}[!t]
\caption{The impact of mask ratio on domain projection.}
\centering
\fontsize{2.6}{3}\selectfont\rmfamily
\resizebox{0.8\columnwidth}{!}{
\begin{tabular}{ccc}
\hline
\multicolumn{3}{c}{MMIOC-1M Open Scene (RTVPNet-L)}   \\\hline
Mask ratio           & AP             & AP50          \\\hline
10\%                 & 11.0           & 22.1          \\
13\%                 & 13.2           & 24.7          \\
15\%                 & 11.4           & 24.5          \\
17\%                 & 13.6           & 26.8          \\
\rowcolor[HTML]{E6E0E0} 
20\%                 & 17.4           & 30.7          \\
23\%                 & 14.2           & 26.5          \\
30\%                 & 12.9           & 23.3          \\\hline
\multicolumn{3}{c}{MMIOC-1M Closed Scene (RTVPNet-S)} \\\hline
15\%                 & 30.5           & 53.2          \\
17\%                 & 32.0           & 55.8          \\
\rowcolor[HTML]{E6E0E0} 
20\%                 & 36.7           & 60.7          \\
23\%                 & 33.8           & 57.4         \\\hline
\end{tabular}
\label{Domainmask}}
\end{table}

\subsubsection{Refined Visual Prompts} 

As described in section \ref{Refined Visual Prompt} regarding motivation, we propose Refined Visual Prompts to obtain fine-grained object-related visual prompts by energy-activating the coarse-grained features and sparse sampling mechanism. We study the effects of different feature activation methods on MMIOC-1M. Table \ref{attentioncomp} shows that the convolutional attention-based model struggles to achieve good results because it is prone to local features and noise redundancy. Its fixed receptive field is difficult to adapt to the dynamic feature extraction of multi-scale objects. Due to global weight dispersion, self-attention leads to insufficient discriminative power of fine-grained features. In contrast, our method significantly improves the discriminative power of fine-grained visual prompts by establishing an energy field of global features. In particular, under the synergistic effect of the sparse sampling mechanism, energy activation can effectively suppress background noise interference and focus visual prompts on object-related features.

\begin{table}[!t]
\caption{Effects of different activation methods on Refined Visual Prompts.}
\centering
\fontsize{2.6}{3}\selectfont\rmfamily
\resizebox{0.9\columnwidth}{!}{
\begin{tabular}{ccc}
\hline
\multicolumn{3}{c}{MMIOC-1M Open Scene (RTVPNet-S)}   \\\hline
Activate method                & AP        & AP50     \\\hline
ECA \cite{eca}                           & 12.1      & 23.5     \\
CBAM \cite{woo2018cbam}                           & 13.8      & 24.9     \\
MCA \cite{2023mca}                           & 14.0      & 25.6     \\
Self-Attn \cite{liu2021swin}                     & 13.7      & 24.2     \\\hline
\rowcolor[HTML]{E6E0E0} 
Energy Activation (Ours)       & 14.9      & 26.7     \\\hline
\multicolumn{3}{c}{MMIOC-1M Closed Scene (RTVPNet-L)} \\\hline
ECA                            & 36.4      & 59.5     \\
CBAM                           & 37.1      & 61.4     \\
MCA                            & 38.9      & 62.6     \\
Self-Attn                      & 39.9      & 63.0     \\\hline
\rowcolor[HTML]{E6E0E0} 
Energy Activation (Ours)       & 41.3      & 64.9    \\\hline
\end{tabular}
\label{attentioncomp}}
\end{table}

The sparse sampling mechanism uses different size patches to select object-related high-frequency features. As shown in Table \ref{spapatch}, the patch size significantly affects the accuracy. This is because the sparsity of industrial defects leads to obvious high-frequency features of defects. The smaller the patch, the more conducive it is to narrow the retrieval range and make it easier to find high-frequency features. However, too small patches will increase the cost of sparse sampling, so we choose a patch size of 32.

\begin{table}[!t]
\caption{Effects of different patch sizes on Refined Visual Prompts.}
\centering
\fontsize{2.6}{3}\selectfont\rmfamily
\resizebox{0.8\columnwidth}{!}{
\begin{tabular}{ccc}
\hline
\multicolumn{3}{c}{MMIOC-1M Open Scene (RTVPNet-L)}   \\\hline
Patch Size           & AP             & AP50          \\\hline
\rowcolor[HTML]{E6E0E0} 
32                   & 17.4           & 30.7          \\
64                   & 15.4           & 25.8          \\
128                  & 13.2           & 22.7          \\\hline
\multicolumn{3}{c}{MMIOC-1M Closed Scene (RTVPNet-S)} \\\hline
\rowcolor[HTML]{E6E0E0} 
32                   & 36.7           & 60.7          \\
64                   & 35.8           & 59.4          \\
128                  & 34.3           & 57.0         \\\hline
\end{tabular}
\label{spapatch}}
\end{table}

\begin{figure}[!h]
\centering
\includegraphics[width=2.8in]{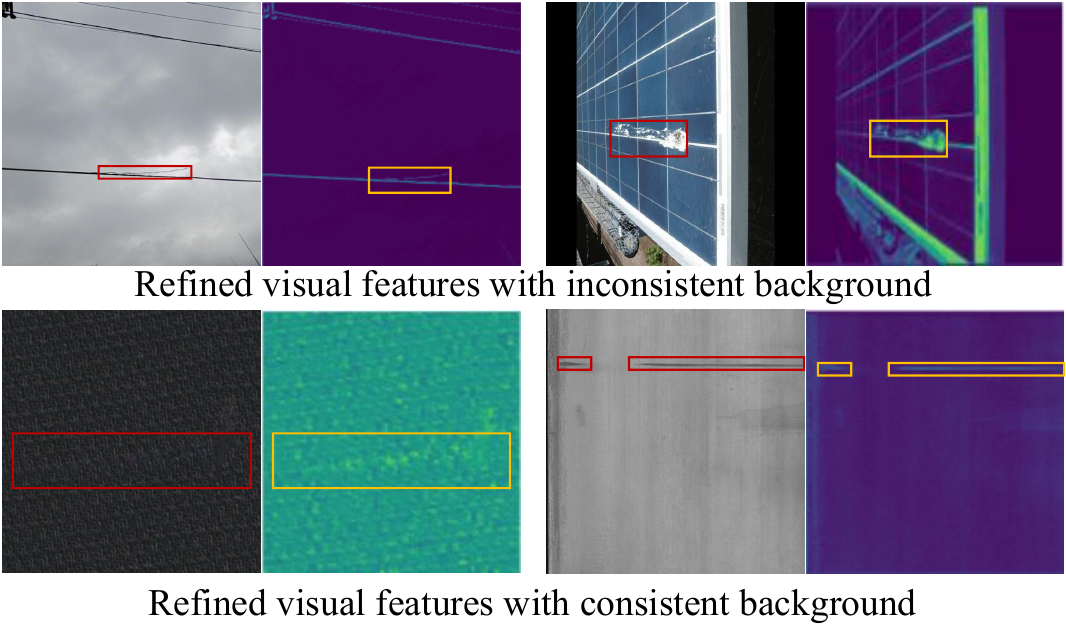}
\caption{Refined visual prompts visualization. The yellow area represents a higher degree of attention. The red box represents the location of the defect. The refined visual prompts can overcome the interference of background diversity.}
\label{Heat}
\end{figure}

To further analyze the effect of Refined Visual Prompts, we visualize the Refined Visual Prompts. Fig.~\ref{Heat} shows the effect of visualization. For defects with inconsistent backgrounds (surfaces with messy textures or areas with sudden changes in illumination), the refined visual prompts can still be distinguished from the complex background noise. This is because the energy field modeling can accurately find the edge of the defect, but some similar background features will still be noticed. For defect with consistent backgrounds, the features in the yellow box show higher attention. The sparse sampling mechanism effectively suppresses the interference of homogeneous backgrounds. In summary, the results confirm that the RTVPNet can overcome the interference of background diversity and enhance the sensitivity of subtle defects in multiple industrial scenarios, providing highly robust visual prompts for complex industrial environments.

\subsubsection{Visual-Text Prompt Bidirectional Interaction} 
As described in section \ref{Visual-Text Mutual}, \ref{Text-Visual Mutual} regarding motivation, we propose Visual Text prompt bidirectional interaction achieves complementary enhancement of text and visual prompts. Table \ref{VisualText} verifies the effectiveness of the visual-text prompt bidirectional interaction. It can be seen that the bidirectional visual and text prompt interaction has a more effective result. The experiment fully demonstrates that the Visual-Text prompt bidirectional interaction overcomes the limitations of single-modality prompts. We explicitly align with and interacts with the text and visual embedding space, enabling the conversion of visual prompts into textual. The text prompt enhances adaptability to visual details through iterative interaction.

\begin{table}[!t]
\caption{The impact of Visual-Text prompt bidirectional interaction.}
\centering
\fontsize{2.6}{3}\selectfont\rmfamily
\resizebox{0.8\columnwidth}{!}{
\begin{tabular}{ccc}
\hline
\multicolumn{3}{c}{MMIOC-1M Open Scene (RTVPNet-L)}  \\ \hline
Method                   & AP                & AP50            \\ \hline
Visual-Text              & 13.3              & 25.2            \\ 
Text-Visual              & 13.7              & 25.7            \\ 
\rowcolor[HTML]{E6E0E0} 
Both                     & 17.4              & 30.7            \\ \hline
\multicolumn{3}{c}{MMIOC-1M Closed Scene (RTVPNet-S)} \\ \hline
Visual-Text              & 31.3              & 53.7            \\ 
Text-Visual              & 31.6              & 53.2            \\ 
\rowcolor[HTML]{E6E0E0} 
Both                     & 36.7              & 60.7            \\ \hline
\end{tabular}
\label{VisualText}}
\end{table}

\begin{figure}[!h]
\centering
\includegraphics[width=2.5in]{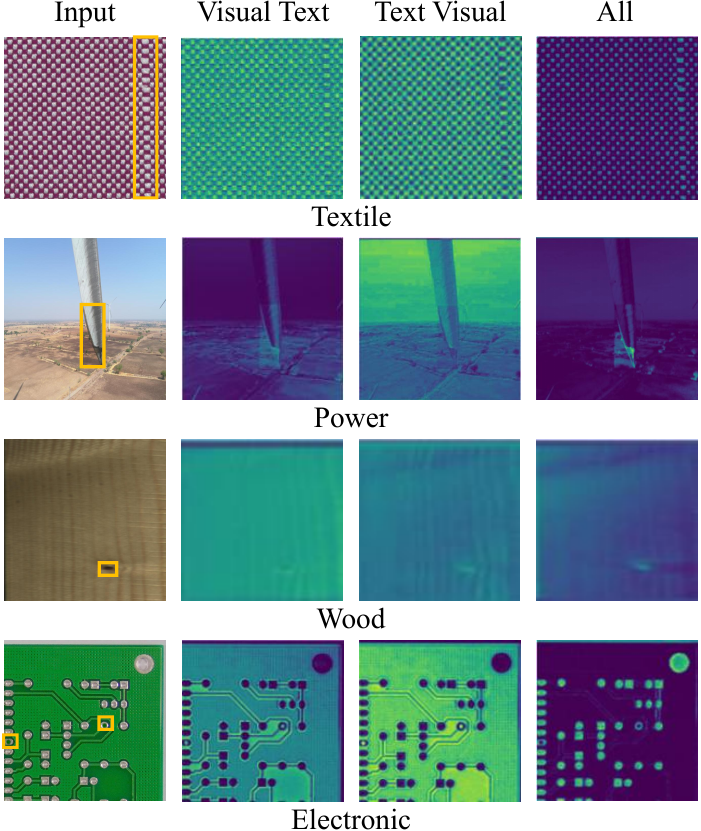}
\caption{Visualization of the Visual-Text prompt bidirectional interaction. The yellow box in the left column shows the object. We select objects with single background and complex background. It can be seen that our method is conducive to querying the semantic information and suppressing the background noise.}
\label{Fearureinter}
\end{figure}

Fig.~\ref{Fearureinter} verifies the significant advantages of the Visual-Text prompt bidirectional interaction in multi-modal feature extraction and background noise suppression. From the middle column, Although the pure text guidance (Visual-Text) can locate the target, the texture details are blurred, and the residue of background noise is obvious. From the right column, the pure visual guidance (Text Visual) causes the solder joint contour to diverge due to the lack of semantic constraints in common object scenes. The bidirectional interactive results (All) are enhanced by the complementary enhancement of text and visual prompts, which significantly suppresses background interference while retaining the key features of the target area. This breakthrough stems from the explicit alignment and interaction of the cross-modal prompt embedding space. The text prompt strengthens the target abstract attributes through semantic constraints, and the visual prompt iteratively corrects the local details.

\subsubsection{Evaluation on Different Visual Encoders and LVLMs} 
We ablate large visual models of various sizes to verify the visual prompt effects. The experimental results are shown in the Table \ref{visual_encoders}. With the parameters increase of model, AP increases slightly. However, a larger ViT will also lead to a decrease in AP, making it difficult for the model to migrate to the target domain. However, the increase in ViT will also lead to a computational burden for the model. Therefore, we choose Mobile-SAM Tiny. It is worth noting that the last two rows in the table demonstrate the effectiveness of RTVP in improving the performance of LVLMs.

\begin{table}
\caption{Comparison of different visual encoders in MMIOC-1M open scenarios.}
\fontsize{4}{5}\selectfont\rmfamily
\centering
\resizebox{1\columnwidth}{!}{
\begin{tabular}{ccccc}
\hline
Method    & AP   & AP50 & AP75 & Parameters \\\hline
SAM-B \cite{sam}    & 13.8 & 25.3 & 12.0 & 752        \\
Fast-SAM \cite{fastsam}  & 16.4 & 27.9 & 15.1 & 201        \\
Edge-SAM  & 11.1 & 22.8 & 10.6 & 148        \\
\rowcolor[HTML]{E6E0E0}
RTVPNet-S & 14.9 & 26.7 & 13.6 & 76        \\\hline 
Qwen3-VL 4B \cite{2025Qwen3} & 14.6 & 27.2 & 12.7 & -        \\\hline
\rowcolor[HTML]{EFEFEF}
Qwen3-VL 4B+RTVP & 15.1 & 27.6 & 13.3 & -\\\hline
\end{tabular}}

\label{visual_encoders}
\end{table}

\subsubsection{Impact of Different Pre-train Data Amount} 

MMIOC-1M is divided into closed and open scene tasks. The closed-scene task contains all the training set data of the open scene. We investigated the impact of different pre-training data sizes on the open-scene task. Experimental results for the RTVPNet-S are shown in Table \ref{diff_data}. The results show that using larger-scale pre-training data has a negative gain for open-scene tasks.  Because there is no semantic association between the additional data and the validation set, it is easy for the model to reduce its recognition ability for invisible categories. Therefore, objects related to the semantics of invisible classes are set in the open scene of MMIOC-1M for training (for example, dirty glass$\to$dirty steel, aluminum holes$\to$PCB gaps).

Furthermore, we investigated the impact of different pre-training data sizes on detection accuracy. We evenly divided MMIOC-1M into subsets of varying sizes by category. The results in Table \ref{diff_data} for the closed-scene task show that the size of the pre-training data used is directly proportional to the accuracy. Therefore, large-scale pre-training is essential. The experiments fully demonstrate the effectiveness of MMIOC-1M in promoting large-scale data pre-training.
\begin{table}
\caption{The impact of different amounts of pre-training data in MMIOC-1M open scenarios.}
\centering
\resizebox{1\columnwidth}{!}{
\begin{tabular}{cccccccccc}
\hline
\multicolumn{10}{c}{Pre-training data results in open scenarios}                                                                                                                                                                                         \\ \hline
\multicolumn{2}{c}{Pretrain data}                                         & \multicolumn{3}{c}{AP}                                          & \multicolumn{3}{c}{AP50}                                             & \multicolumn{2}{c}{AP75}            \\ \hline
\multicolumn{2}{c}{MMIOC-1M (351 Classes)}                                & \multicolumn{3}{c}{9.7}                                         & \multicolumn{3}{c}{18.6}                                             & \multicolumn{2}{c}{7.8}             \\
\multicolumn{2}{c}{Original (94 train classes and 64 val classes)}        & \multicolumn{3}{c}{14.9}                                        & \multicolumn{3}{c}{26.7}                                             & \multicolumn{2}{c}{13.6}            \\ \hline
\multicolumn{10}{c}{Pre-training data results in closed scenarios}                                                                                                                                                                                       \\ \hline
\multicolumn{1}{c|}{Method}                    & \multicolumn{3}{c|}{RTVPNet-S}                                         & \multicolumn{3}{c|}{YOLOWorld-S}                                       & \multicolumn{3}{c}{YOLOv12S}                          \\ \hline
\multicolumn{1}{c|}{Ratio}                     & \multicolumn{2}{c}{AP}            & \multicolumn{1}{c|}{AP50}          & \multicolumn{2}{c}{AP}            & \multicolumn{1}{c|}{AP50}          & \multicolumn{2}{c}{AP}            & AP50              \\ \hline
\multicolumn{1}{c|}{0.1M}                      & \multicolumn{2}{c}{10.8}          & \multicolumn{1}{c|}{21.2}          & \multicolumn{2}{c}{6.3}           & \multicolumn{1}{c|}{13.3}          & \multicolumn{2}{c}{5.4}           & 11.7              \\
\multicolumn{1}{c|}{0.5M}                      & \multicolumn{2}{c}{19.6}          & \multicolumn{1}{c|}{28.1}          & \multicolumn{2}{c}{10.4}          & \multicolumn{1}{c|}{21.0}          & \multicolumn{2}{c}{12.7}          & 23.6              \\
\multicolumn{1}{c|}{1M}                        & \multicolumn{2}{c}{\textbf{36.7}} & \multicolumn{1}{c|}{\textbf{60.7}} & \multicolumn{2}{c}{\textbf{20.4}} & \multicolumn{1}{c|}{\textbf{38.1}} & \multicolumn{2}{c}{\textbf{23.5}} & \textbf{47.8}     \\ \hline
\end{tabular}}
\label{diff_data}
\end{table}

\section{Discussion}
\label{sec_Discussion}

The experiments demonstrate the universality of various multi-modal task features learned from MMIOC-1M, indicating the value of MMIOC-1M. Below, we discusses some potential research issues and methods brought by MMIOC-1M.

\noindent(1) \textbf{Research on robust identification of industrial defect detection:} Current defect detection methods perform well in traditional datasets, but not well in large-scale open industrial defect detection scenarios such as MMIOC-1M. The reason is that the diversity and scale of industrial data make the visual patterns of defects complex, which is difficult for existing methods to cope with. We explore technologies such as refined visual prompts and achieves good results on the MMIOC-1M. In the future, the detection performance can be further improved with the help of LVLMs.

\noindent(2) \textbf{Research on multi-task extension of MMIOC-1M:} We encourage the application of models trained on the MMIOC-1M dataset to a wider range of industrial tasks and promotes the evolution and expansion of datasets. By incorporating rich attribute annotations, performing region-level and pixel-level anomaly annotations, etc., the application scenarios can be broadened, such as introducing semantic segmentation labels to support complex tasks. The multi-modal characteristics of MMIOC-1M combined with the multi-modal interaction capabilities of RTVPNet can improve the performance of the model in multi-modal tasks and enhance the accuracy, robustness and adaptability of detection.

\section{Conclusion}
\label{sec_Conclusion}
The significant differences between industrial and natural scenes make the applicationof LVLMs challenging. To address the problem of large-scale data scarcity in the industry, we present the first Multi-Modal Industrial Open-Closed benchmark (MMIOC-1M). MMIOC-1M supports both open and closed scene tasks, featuring a rich array of industry categories, including 14 super categories, 29 scenes, and 351 subcategories. In addition, MMIOC-1M is the largest multi-modal industrial defect benchmark and the first paradigm extended to industrial open-scene defect detection. We believe that this will help develop large-scale industrial expert LVLMs and also enable researchers to utilize MMIOC-1M for future research on industrial detection-related tasks, such as cross-scene and cross-super-class transfer learning. Therefore, MMIOC-1M can serve as a new benchmark for industrial open and closed scenes. Based on MMIOC-1M, we provide a Refined Text-Visual Prompt Network (RTVPNet) for industrial defect detection tasks. RTVPNet promotes transfer learning of Mobile-Segment Anything in industrial scenarios through expert-guided domain transfer. Secondly, RTVPNet proposes Refined Visual Prompts and text-visual interaction methods to promote cross-modal mutual matching and understanding. Experiments have demonstrated the effectiveness of RTVPNet in addressing challenges such as inter-class similarity, intra-class difference, and long-tail distribution of MMIOC-1M. In the future, we believe that more baselines based on MMIOC-1M will emerge.

\section{Acknowledgement}
Portions of this work were previously presented at AAAI 2025 under the title “Zero-Shot Learning in Industrial Scenarios: New Large-Scale Benchmark, Challenges and Baseline”. This work was supported by the joint funds of the National Natural Science Foundation of China under Grant U24A20221, Key R D Program of Shandong Province of China under Grant 2023CXGC010112, Distinguished Young Scholar of Shandong Province under Grant ZR2023JQ025, Taishan Scholars Program under Grant tstp20250708, Major Basic Research Projects of Shandong Province under Grant ZR2022ZD32, national funds through FCT (Fundação para a Ciência e a Tecnologia), under the project - UID/04152/2025 - Centro de Investigação em Gestão de Informação (MagIC)/NOVA IMS and UID/PRR/04152/2025.

\bibliographystyle{IEEEtran}
\bibliography{name.bib}

@article{2021YOLOX,
  title={YOLOX: Exceeding YOLO Series in 2021},
  author={Ge, Z. and Liu, S.},
  journal={arXiv preprint arXiv:2107.08430},
  year={2021},
}

@inproceedings{zheng2020distance,
  title={Distance-IoU loss: Faster and better learning for bounding box regression},
  author={Zheng, Zhaohui and Wang, Ping},
  booktitle={AAAI},
  volume={34},
  number={07},
  pages={12993--13000},
  year={2020}
}

@article{li2022yolov6,
  title={YOLOv6: A single-stage object detection framework for industrial applications},
  author={Li, Chuyi and Li, Lulu},
  journal={arXiv preprint arXiv:2209.02976},
  year={2022}
}

@inproceedings{resnet,
  title={Deep residual learning for image recognition},
  author={He, Kaiming and Zhang, Xiangyu},
  booktitle={CVPR},
  pages={770--778},
  year={2016}
}

@InProceedings{resnext,
author = {Xie, Saining and Girshick, Ross},
title = {Aggregated Residual Transformations for Deep Neural Networks},
booktitle = {CVPR},
month = {July},
year = {2017}
}

@inproceedings{liu2021swin,
  title={Swin transformer: Hierarchical vision transformer using shifted windows},
  author={Liu, Ze and Lin, Yutong},
  booktitle={ICCV},
  pages={10012--10022},
  year={2021}
}

@inproceedings{COCO,
Author = {Lin, Tsung-Yi and Maire, Michael},
Editor = {Fleet, D and Pajdla, T and Schiele, B and Tuytelaars, T},
Title = {Microsoft COCO: Common Objects in Context},
Booktitle = {ECCV},
Year = {2014},
Volume = {8693},
Pages = {740-755},
}

@inproceedings{wang2022yolov7,
  title={YOLOv7: Trainable bag-of-freebies sets new state-of-the-art for real-time object detectors},
  author={Wang, Chien-Yao},
  booktitle={CVPR},
  pages={7464--7475},
  year={2023}
}

@INPROCEEDINGS{shouji,
  author={Zhang, Jian and Ding, Runwei},
  booktitle={ICASSP}, 
  title={FDSNeT: An Accurate Real-Time Surface Defect Segmentation Network}, 
  year={2022},
  volume={},
  number={},
  pages={3803-3807}}

@InProceedings{sam,
    author    = {Kirillov, Alexander and Mintun, Eric},
    title     = {Segment Anything},
    booktitle = {ICCV},
    month     = {October},
    year      = {2023},
    pages     = {4015-4026},
}

@misc{fastsam,
      title={Fast Segment Anything}, 
      author={Xu Zhao and Wenchao Ding},
      year={2023},
      eprint={2306.12156},
      archivePrefix={arXiv},
      primaryClass={cs.CV},
      url={https://arxiv.org/abs/2306.12156}, 
}

@article{fastersam,
  title={Faster segment anything: Towards lightweight sam for mobile applications},
  author={Zhang, Chaoning},
  journal={arXiv preprint arXiv:2306.14289},
  year={2023}
}

@article{matcher,
  title={Matcher: Segment anything with one shot using all-purpose feature matching},
  author={Liu, Yang and Zhu, Muzhi},
  journal={arXiv preprint arXiv:2305.13310},
  year={2023}
}

@article{personalize,
  title={Personalize segment anything model with one shot},
  author={Zhang, Renrui and Jiang, Zhengkai},
  journal={arXiv preprint arXiv:2305.03048},
  year={2023}
}

@article{yao2024cpt,
  title={Cpt: Colorful prompt tuning for pre-trained vision-language models},
  author={Yao, Yuan and Zhang, Ao},
  journal={AI Open},
  volume={5},
  pages={30--38},
  year={2024},
  publisher={Elsevier}
}

@article{subramanian2022reclip,
  title={Reclip: A strong zero-shot baseline for referring expression comprehension},
  author={Subramanian, Sanjay},
  journal={arXiv preprint arXiv:2204.05991},
  year={2022}
}

@article{yan2023cocoopter,
  title={CoCoOpter: Pre-train, prompt, and fine-tune the vision-language model for few-shot image classification},
  author={Yan, Jie and Xie, Yuxiang},
  journal={Int. J. Multi. Inform. Retri.},
  volume={12},
  number={2},
  pages={27},
  year={2023},
  publisher={Springer}
}

@article{hu2023efficiently,
  title={How to efficiently adapt large segmentation model (sam) to medical images},
  author={Hu, Xinrong and Xu, Xiaowei},
  journal={arXiv preprint arXiv:2306.13731},
  year={2023}
}

@article{gpt4,
  title={Gpt-4 technical report},
  author={Achiam, Josh and Adler, Steven},
  journal={arXiv preprint arXiv:2303.08774},
  year={2023}
}

@inproceedings{llava,
  title={Visual instruction tuning},
  author={Liu, Haotian and Li, Chunyuan},
  journal={NeurIPS},
  volume={36},
  year={2024}
}

@inproceedings{clip,
  title={Learning transferable visual models from natural language supervision},
  author={Radford, Alec and Kim, Jong Wook},
  booktitle={ICML},
  pages={8748--8763},
  year={2021},
  organization={PMLR}
}

@inproceedings{li2023blip,
  title={Blip-2: Bootstrapping language-image pre-training with frozen image encoders and large language models},
  author={Li, Junnan and Li, Dongxu},
  booktitle={ICML},
  pages={19730--19742},
  year={2023},
  organization={PMLR}
}

@article{wu2023medicalsam,
    title={Medical sam adapter: Adapting segment anything model for medical image segmentation},
  author={Wu, Junde and Wang, Ziyue},
  journal={Medical image analysis},
  volume={102},
  pages={103547},
  year={2025},
  publisher={Elsevier}
}

@inproceedings{liu2023groundingdino,
  title={Grounding dino: Marrying dino with grounded pre-training for open-set object detection},
  author={Liu, Shilong and Zeng, Zhaoyang},
  booktitle={ECCV},
  pages={38--55},
  year={2024},
  organization={Springer}
}

@article{ahmadi2023application,
  title={Application of segment anything model for civil infrastructure defect assessment},
  author={Ahmadi, Mohsen and Lonbar, Ahmad Gholizadeh},
  journal={arXiv preprint arXiv:2304.12600},
  year={2023}
}

@article{xu2023eviprompt,
  title={Eviprompt: A training-free evidential prompt generation method for segment anything model in medical images},
  author={Xu, Yinsong and Tang, Jiaqi},
  journal={arXiv preprint arXiv:2311.06400},
  year={2023}
}

@article{jie2023adaptershadow,
  title={AdapterShadow: Adapting Segment Anything Model for Shadow Detection},
  author={Jie, Leiping and Zhang, Hui},
  journal={arXiv preprint arXiv:2311.08891},
  year={2023}
}

@article{shin2020autoprompt,
  title={Autoprompt: Eliciting knowledge from language models with automatically generated prompts},
  author={Shin, Taylor and Razeghi, Yasaman},
  journal={arXiv preprint arXiv:2010.15980},
  year={2020}
}

@article{yang2024fine,
  title={Fine-grained visual prompting},
  author={Yang, Lingfeng and Wang, Yueze},
  journal={NeurIPS},
  volume={36},
  year={2024}
}

@inproceedings{sun2024vrp,
  title={VRP-SAM: SAM with visual reference prompt},
  author={Sun, Yanpeng and Chen, Jiahui},
  booktitle={CVPR},
  pages={23565--23574},
  year={2024}
}

@inproceedings{glip,
  title={Grounded language-image pre-training},
  author={Li, Liunian Harold and Zhang, Pengchuan},
  booktitle={CVPR},
  pages={10965--10975},
  year={2022}
}

@inproceedings{yoloworld,
  title={Yolo-world: Real-time open-vocabulary object detection},
  author={Cheng, Tianheng and Song, Lin},
  booktitle={CVPR},
  pages={16901--16911},
  year={2024}
}

@inproceedings{bergmann2019mvtec,
  title={MVTec AD--A comprehensive real-world dataset for unsupervised anomaly detection},
  author={Bergmann, Paul and Fauser, Michael},
  booktitle={CVPR},
  pages={9592--9600},
  year={2019}
}

@inproceedings{he2022mae,
  title={Masked autoencoders are scalable vision learners},
  author={He, Kaiming and Chen, Xinlei},
  booktitle={CVPR},
  pages={16000--16009},
  year={2022}
}

@article{YOLOv8,
  title={YOLOv8},
  author={ultralytics},
  journal={[Online]. Available: https://github.com/ultralytics/yolov8},
  year={2023},
}

@article{wang2024yolov10,
  title={Yolov10: Real-time end-to-end object detection},
  author={Wang, Ao and Chen, Hui},
  journal={NeurIPS},
  volume={37},
  pages={107984--108011},
  year={2024}
}

@inproceedings{kamath2021mdetr,
  title={Mdetr-modulated detection for end-to-end multi-modal understanding},
  author={Kamath, Aishwarya and Singh, Mannat},
  booktitle={CVPR},
  pages={1780--1790},
  year={2021}
}

@inproceedings{shao2019objects365,
  title={Objects365: A large-scale, high-quality dataset for object detection},
  author={Shao, Shuai and Li, Zeming},
  booktitle={ICCV},
  pages={8430--8439},
  year={2019}
}

@article{gu2021open,
  title={Open-vocabulary object detection via vision and language knowledge distillation},
  author={Gu, Xiuye and Lin, Tsung-Yi},
  journal={arXiv preprint arXiv:2104.13921},
  year={2021}
}

@article{t-sne,
  title={Visualizing data using t-SNE.},
  author={Van der Maaten, Laurens and Hinton, Geoffrey},
  journal={J. mach. lear. resea.},
  volume={9},
  number={11},
  year={2008}
}

@inproceedings{2021VT,
  title={VT-ADL: A vision transformer network for image anomaly detection and localization},
  author={Mishra, Pankaj and Verk, Riccardo},
  booktitle={ISIE},
  pages={01--06},
  year={2021},
  organization={IEEE}
}

@article{2020Deep,
  title={Deep learning for medical anomaly detection--a survey},
  author={Fernando, Tharindu and Gammulle, Harshala},
  journal={ACM Computing Surveys (CSUR)},
  volume={54},
  number={7},
  pages={1--37},
  year={2021},
  publisher={ACM New York, NY, USA}
}

@article{0Pixel,
  title={Pixel-Level Contrastive Pretrainer for Industrial Image Representation},
  author={Zhu, Bingke and Chen, Yingying},
  journal={IEEE Trans. on Instru. and Measure.},
  volume={73},
  year={2024},
}

@ARTICLE{lite-yolo,
  author={Li, Dahua and Lu, Yang},
  journal={IEEE Trans. on Instru. and Measure.}, 
  title={LiteYOLO-ID: A Lightweight Object Detection Network for Insulator Defect Detection}, 
  year={2024},
  volume={73},
  number={},
  pages={1-12},
  doi={10.1109/TIM.2024.3418082}}

@article{lf,
  title={LF-YOLO: A lighter and faster yolo for weld defect detection of X-ray image},
  author={Liu, Moyun and Chen, Youping},
  journal={IEEE Sens. J.},
  volume={23},
  number={7},
  pages={7430--7439},
  year={2023},
  publisher={IEEE}
}

@article{InfoNce,
  title={Representation learning with contrastive predictive coding},
  author={Oord, Aaron van den and Li, Yazhe},
  journal={arXiv preprint arXiv:1807.03748},
  year={2018}
}

@article{2025DeepSeek,
  title={Deepseek-r1: Incentivizing reasoning capability in llms via reinforcement learning},
  author={Guo, Daya and Yang, Dejian},
  journal={arXiv preprint arXiv:2501.12948},
  year={2025}
}

@inproceedings{Defect_spectrum,
  title={Defect spectrum: a granular look of large-scale defect datasets with rich semantics},
  author={Yang, Shuai and Chen, Zhifei},
  booktitle={ECCV},
  pages={187--203},
  year={2024},
  organization={Springer}
}

@article{bai2023vision,
  title={Vision datasets: A benchmark for vision-based industrial inspection},
  author={Bai, Haoping and Mou, Shancong},
  journal={arXiv preprint arXiv:2306.07890},
  year={2023}
}

@article{jiang2024mmad,
  title={Mmad: The first-ever comprehensive benchmark for multimodal large language models in industrial anomaly detection},
  author={Jiang, Xi and Li, Jian},
  journal={arXiv preprint arXiv:2410.09453},
  year={2024}
}

@inproceedings{Visa,
  title={Spot-the-difference self-supervised pre-training for anomaly detection and segmentation},
  author={Zou, Yang and Jeong, Jongheon},
  booktitle={ECCV},
  pages={392--408},
  year={2022},
  organization={Springer}
}

@inproceedings{yang20253cad,
  title={3CAD: A Large-Scale Real-World 3C Product Dataset for Unsupervised Anomaly Detection},
  author={Yang, Enquan and Xing, Peng},
  booktitle={AAAI},
  volume={39},
  number={9},
  pages={9175--9183},
  year={2025}
}

@article{zhang2024pkugood,
  title={PKU-GoodsAD: A supermarket goods dataset for unsupervised anomaly detection and segmentation},
  author={Zhang, Jian and Ding, Runwei},
  journal={IEEE Robot. and Auto. Lett.},
  volume={9},
  number={3},
  pages={2008--2015},
  year={2024},
  publisher={IEEE}
}

@inproceedings{wang2024real,
  title={Real-iad: A real-world multi-view dataset for benchmarking versatile industrial anomaly detection},
  author={Wang, Chengjie and Zhu, Wenbing},
  booktitle={CVPR},
  pages={22883--22892},
  year={2024}
}

@inproceedings{li2024mulsen,
   title={Multi-sensor object anomaly detection: Unifying appearance, geometry, and internal properties},
  author={Li, Wenqiao and Zheng, Bozhong},
  booktitle={CVPR},
  pages={9984--9993},
  year={2025}
}

@article{IndustrialTextileDataset,
  title={Distillation-based fabric anomaly detection},
  author={Thomine, Simon and Snoussi, Hichem},
  journal={Textile Research Journal},
  volume={94},
  number={5-6},
  pages={552--565},
  year={2024},
}

@article{liu2024yoloUNI,
  title={YOLO-UniOW: Efficient Universal Open-World Object Detection},
  author={Liu, Lihao and Feng, Juexiao},
  journal={arXiv preprint arXiv:2412.20645},
  year={2024}
}

@article{chaudhuri2024relational,
  title={Relational Proxies: Fine-Grained Relationships as Zero-Shot Discriminators},
  author={Chaudhuri, Abhra and Mancini, Massimiliano},
  journal={IEEE Trans. on Pattern Anal. and Mach. Intell.},
  year={2024},
  publisher={IEEE}
}

@inproceedings{Zero-shotLearning, 
title={Zero-Shot Learning in Industrial Scenarios: New Large-Scale Benchmark, Challenges and Baseline}, 
volume={39}, 
url={https://ojs.aaai.org/index.php/AAAI/article/view/33124}, 
DOI={10.1609/aaai.v39i10.33124}, 
number={10}, 
booktitle={AAAI}, 
author={Zhang, Zekai and Chen, Qinghui}, 
year={2025},
month={Apr.}, 
pages={10357-10366}}

@misc{yolov11,
      title={YOLOv11: An Overview of the Key Architectural Enhancements}, 
      author={Rahima Khanam and Muhammad Hussain},
      year={2024},
      eprint={2410.17725},
      archivePrefix={arXiv},
      primaryClass={cs.CV},
      url={https://arxiv.org/abs/2410.17725}, 
}

@article{LVIS,
  author       = {Agrim Gupta and
                  Piotr Doll{\'{a}}r},
  title        = {{LVIS:} {A} Dataset for Large Vocabulary Instance Segmentation},
  journal      = {CoRR},
  volume       = {abs/1908.03195},
  year         = {2019},
  bibsource    = {dblp computer science bibliography, https://dblp.org}
}

@INPROCEEDINGS{Mamba-YOLO-World,
  author={Wang, Haoxuan and He, Qingdong},
  booktitle={ICASSP}, 
  title={Mamba-YOLO-World: Marrying YOLO-World with Mamba for Open-Vocabulary Detection}, 
  year={2025},
  volume={},
  number={},
  pages={1-5},
  }

@misc{DOSOD,
      title={A Light-Weight Framework for Open-Set Object Detection with Decoupled Feature Alignment in Joint Space}, 
      author={Yonghao He and Hu Su},
      year={2024},
      eprint={2412.14680},
      archivePrefix={arXiv},
      primaryClass={cs.CV},
      url={https://arxiv.org/abs/2412.14680}, 
}

@inproceedings{eca,
  title={ECA-Net: Efficient channel attention for deep convolutional neural networks},
  author={Wang, Qilong and Wu, Banggu},
  booktitle={CVPR},
  pages={11534--11542},
  year={2020}
}

@inproceedings{woo2018cbam,
  title={Cbam: Convolutional block attention module},
  author={Woo, Sanghyun and Park, Jongchan},
  booktitle={ECCV},
  pages={3--19},
  year={2018}
}

@article{2023mca,
    title={MCA: Multidimensional collaborative attention in deep convolutional neural networks for image recognition},
    author={Yu, Yang and Zhang, Yi},
    journal={Eng. App. Arti. Intel.},
    volume={126},
    pages={107079},
    year={2023},
    publisher={Elsevier}
}

@article{tian2025yolov12,
  title={Yolov12: Attention-centric real-time object detectors},
  author={Tian, Yunjie and Ye, Qixiang},
  journal={arXiv preprint arXiv:2502.12524},
  year={2025}
}

@article{wang2024mambayolo,
  title={Mamba YOLO: SSMs-based YOLO for object detection},
  author={Wang, Zeyu and Li, Chen},
  journal={arXiv preprint arXiv:2406.05835},
  year={2024}
}

@article{feng2024hyper,
  title={Hyper-yolo: When visual object detection meets hypergraph computation},
  author={Feng, Yifan and Huang, Jiangang},
  journal={IEEE Trans. on Pattern Anal. and Mach. Intell.},
  year={2024},
  publisher={IEEE}
}

@article{YOLO-PCB,
  title={A deep context learning based PCB defect detection model with anomalous trend alarming system},
  author={Lim, JiaYou and Lim, JunYi},
  journal={Results in Engineering},
  volume={17},
  pages={100968},
  year={2023},
  publisher={Elsevier}
}

@article{Etdnet,
  title={ETDNet: Efficient transformer-based detection network for surface defect detection},
  author={Zhou, Hantao and Yang, Rui},
  journal={IEEE Trans. on Inst. and Meas.},
  volume={72},
  pages={1--14},
  year={2023},
  publisher={IEEE}
}

@article{huang2024ssa,
  title={SSA-YOLO: an improved YOLO for hot-rolled strip steel surface defect detection},
  author={Huang, Xiaohua and Zhu, Jiahao},
  journal={IEEE Trans. on Inst. and Meas.},
  year={2024},
  publisher={IEEE}
}

@article{Segment-Not-Perfect,
      author={Ji, Wei and Li, Jingjing},
      journal={Machine Intelligence Research},
      title={Segment Anything Is Not Always Perfect: An Investigation of SAM on Different Real-world Applications},
      year={2024},
      volume={21},
      pages={617--630},
      publisher={Springer}
}

@article{2024LLaVA-ov,
  title={LLaVA-OneVision: Easy Visual Task Transfer},
  author={ Li, Bo  and  Zhang, Yuanhan},
  year={2024},
  url={arXiv preprint arXiv:2408.03326.},
}

@article{2025Qwen3,
  title={Qwen3-Omni Technical Report},
  author={ Xu, Jin  and  Guo, Zhifang},
  year={2025},
  url={arXiv preprint arXiv:2505.09388.},
}

@article{2025Qwen2,
  title={Qwen2.5-VL Technical Report},
  author={ Bai, Shuai  and  Chen, Keqin},
  year={2025},
  url={arXiv preprint arXiv:2502.13923.},
}

@article{Llava-next,
  title={Llava-next-interleave: Tackling multi-image, video, and 3d in large multimodal models},
  author={Li, F.},
  year={2025},
  url={arXiv preprint arXiv:2407.07895.},
}

@ARTICLE{defectglm,
  author={Wang, Huan and Li, Chenxi},
  journal={IEEE Trans. on Indus. Infor.}, 
  title={Large-Scale Visual Language Model Boosted by Contrast Domain Adaptation for Intelligent Industrial Visual Monitoring}, 
  year={2024},
  volume={20},
  number={12},
  pages={14114-14123},
  doi={10.1109/TII.2024.3441638}}

@article{1,
  title={S2DBFT: Spectral-spatial dual-branch fusion transformer for hyperspectral image classification},
  author={Zhang, Yiheng and Wang, Ziqiang and Huang, Meng and Li, Ming and Zhang, Jian and Wang, Shandong and Zhang, Jinglin and Zhang, Heng},
  journal={IEEE Transactions on Geoscience and Remote Sensing},
  year={2025},
  publisher={IEEE}
}

@inproceedings{2,
  title={Implementation of motion estimation based on heterogeneous parallel computing system with OpenCL},
  author={Zhang, Jinglin and Nezan, Jean-Francois and Cousin, Jean-Gabriel},
  booktitle={2012 IEEE 14th International Conference on High Performance Computing and Communication \& 2012 IEEE 9th International Conference on Embedded Software and Systems},
  pages={41--45},
  year={2012},
  organization={IEEE}
}

@article{3,
  title={3D octave and 2D vanilla mixed convolutional neural network for hyperspectral image classification with limited samples},
  author={Feng, Yuchao and Zheng, Jianwei and Qin, Mengjie and Bai, Cong and Zhang, Jinglin},
  journal={Remote Sensing},
  volume={13},
  number={21},
  pages={4407},
  year={2021},
  publisher={MDPI}
}

@article{4,
  title={Learning vertex representations for bipartite networks},
  author={Gao, Ming and He, Xiangnan and Chen, Leihui and Liu, Tingting and Zhang, Jinglin and Zhou, Aoying},
  journal={IEEE transactions on knowledge and data engineering},
  volume={34},
  number={1},
  pages={379--393},
  year={2020},
  publisher={IEEE}
}

@article{5,
  title={Multi-granularity episodic contrastive learning for few-shot learning},
  author={Zhu, Pengfei and Zhu, Zhilin and Wang, Yu and Zhang, Jinglin and Zhao, Shuai},
  journal={Pattern Recognition},
  volume={131},
  pages={108820},
  year={2022},
  publisher={Elsevier}
}

@article{6,
  title={A novel ground-based cloud image segmentation method by using deep transfer learning},
  author={Zhou, Zecheng and Zhang, Feng and Xiao, Haixia and Wang, Fuchang and Hong, Xin and Wu, Kun and Zhang, Jinglin},
  journal={IEEE Geoscience and Remote Sensing Letters},
  volume={19},
  pages={1--5},
  year={2021},
  publisher={IEEE}
}

@article{7,
  title={Ensemble meteorological cloud classification meets internet of dependable and controllable things},
  author={Zhang, Jinglin and Liu, Pu and Zhang, Feng and Iwabuchi, Hironobu and e Ayres, Antonio Artur de H and De Albuquerque, Victor Hugo C and others},
  journal={IEEE Internet of Things Journal},
  volume={8},
  number={5},
  pages={3323--3330},
  year={2020},
  publisher={IEEE}
}

@article{8,
  title={Automated CCA-MWF algorithm for unsupervised identification and removal of EOG artifacts from EEG},
  author={Miao, Minmin and Hu, Wenjun and Xu, Baoguo and Zhang, Jinglin and Rodrigues, Joel JPC and De Albuquerque, Victor Hugo C},
  journal={IEEE Journal of Biomedical and Health Informatics},
  volume={26},
  number={8},
  pages={3607--3617},
  year={2021},
  publisher={IEEE}
}

@article{9,
  title={Supervised learning based discrete hashing for image retrieval},
  author={Ma, Qing and Bai, Cong and Zhang, Jinglin and Liu, Zhi and Chen, Shengyong},
  journal={Pattern Recognition},
  volume={92},
  pages={156--164},
  year={2019},
  publisher={Elsevier}
}

@article{10,
  title={Clothing sale forecasting by a composite GRU--Prophet model with an attention mechanism},
  author={Li, Yuanjiang and Yang, Yi and Zhu, Kai and Zhang, Jinglin},
  journal={IEEE Transactions on Industrial Informatics},
  volume={17},
  number={12},
  pages={8335--8344},
  year={2021},
  publisher={IEEE}
}

@article{11,
  title={Distilled large language model-driven dynamic sparse expert activation mechanism},
  author={Chen, Qinghui and Zhang, Zekai and Zhang, Zaigui and Zhang, Kai and Li, Dagang and Wang, Wenmin and Zhang, Jinglin and Liu, Cong},
  journal={Applied Soft Computing},
  pages={114037},
  year={2025},
  publisher={Elsevier}
}

@article{12,
  title={Dual-path aggregation transformer network for super-resolution with images occlusions and variability},
  author={ Chen, Qinghui  and  Wang, Lunqian  and  Zhang, Zekai  and  Wang, Xinghua  and  Liu, Weilin  and  Xia, Bo  and  Ding, Hao  and  Zhang, Jinglin  and  Xu, Sen  and  Wang, Xin },
  journal={Engineering Applications of Artificial Intelligence},
  volume={139},
  number={PartA},
  year={2025},
}

@inproceedings{13,
  title={KFTD: Koopman-Fourier Time-Differentiable Network for Continuous Ocean Spatiotemporal Forecasting},
  author={Chen, Qinghui and Zhang, Zekai and Liu, Hailong and Zhang, Jinglin and Bai, Cong},
  booktitle={Proceedings of the 32nd ACM SIGKDD Conference on Knowledge Discovery and Data Mining V. 1},
  pages={94--103},
  year={2026}
}

@article{zhang2026novel,
  title={A Novel Dataset and Lightweight Distillation Baseline for Highlight Transparent Object Detection},
  author={Zhang, Zekai and Li, Gang and Zhang, Haijun and Chen, Qinghui and Zhang, Qunshu and Wan, Jin and Xiong, MaoMao and Bai, Cong and Li, Dagang and Zhang, Wenyin and others},
  journal={International Journal of Computer Vision},
  volume={134},
  number={4},
  pages={157},
  year={2026},
  publisher={Springer}
}

@article{zhang2023idd,
  title={IDD-Net: Industrial defect detection method based on Deep-Learning},
  author={Zhang, Zekai and Zhou, Mingle and Wan, Honglin and Li, Min and Li, Gang and Han, Delong},
  journal={Engineering Applications of Artificial Intelligence},
  volume={123},
  pages={106390},
  year={2023},
  publisher={Elsevier}
}

@inproceedings{zhang2025zero,
  title={Zero-shot learning in industrial scenarios: New large-scale benchmark, challenges and baseline},
  author={Zhang, Zekai and Chen, Qinghui and Xiong, Maomao and Ding, Shijiao and Su, Zhanzhi and Yao, Xinjie and Sun, Yiming and Bai, Cong and Zhang, Jinglin},
  booktitle={Proceedings of the AAAI Conference on Artificial Intelligence},
  volume={39},
  number={10},
  pages={10357--10366},
  year={2025}
}

@article{zhang2024representation,
  title={Representation learning based on co-evolutionary combined with probability distribution optimization for precise defect location},
  author={Zhang, Jinglin and Zhang, Zekai and Chen, Qinghui and Li, Gang and Li, Weiyu and Ding, Shijiao and Xiong, Maomao and Zhang, Wenhao and Chen, Shengyong},
  journal={IEEE Transactions on Neural Networks and Learning Systems},
  volume={36},
  number={7},
  pages={11989--12003},
  year={2024},
  publisher={IEEE}
}

@article{zhang2026unification,
  title={Unification of Closed-Open Industrial Detection Scenarios: New Large-Scale Benchmarks, Challenges and Baselines},
  author={Zhang, Zekai and Zhang, Jinglin and Chen, Qinghui and Li, Gang and Chen, Da and Jing, Shuainan and Wang, He and Li, Dagang and Liu, Cong and Bai, Cong and others},
  journal={IEEE Transactions on Pattern Analysis and Machine Intelligence},
  year={2026},
  publisher={IEEE}
}

\vspace{-1cm}
\begin{IEEEbiography}
[{\includegraphics[width=1in,height=1.25in,clip,keepaspectratio]{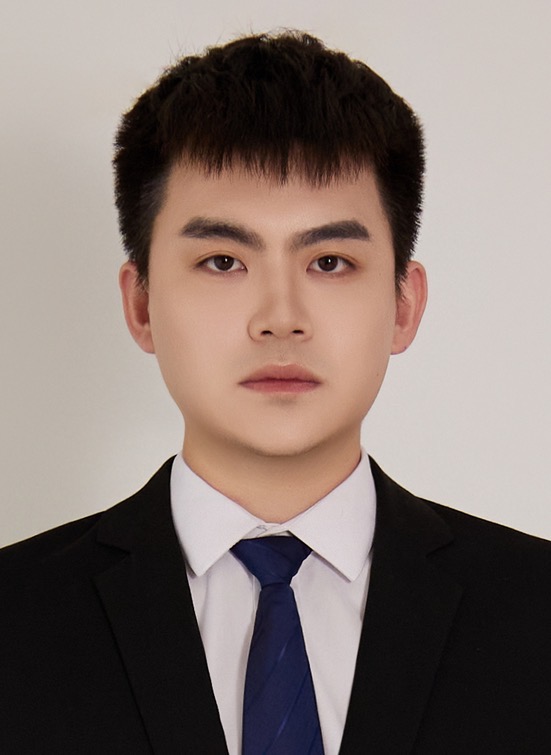}}] {Zekai Zhang} (Student Member, IEEE) received
his master’s degrees from Qilu University of Technology. He is currently pursuing a doctorate degree at the School of Control Science and Engineering, Shandong University. His research interests include computer vision, edge computing, and deep learning.
\end{IEEEbiography}

\vspace{-1cm}
\begin{IEEEbiography}
[{\includegraphics[width=1in,height=1.25in,clip,keepaspectratio]{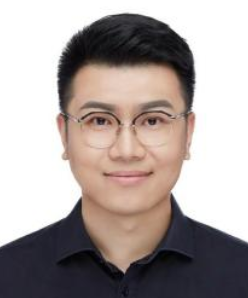}}] {Jinglin Zhang} received the Ph.D. degree in electronics and communication engineering from the National Institute of Applied Sciences, Rennes, France, in 2007, 2010, and 2013, respectively. He is currently a Professor with the School of Control Science and Engineering, Shandong University, Jinan, China. His research interests include computer vision and interdisciplinary research with pattern recognition and atmospheric science. 
\end{IEEEbiography}

\vspace{-1cm}
\begin{IEEEbiography}
[{\includegraphics[width=1in,height=1.25in,clip,keepaspectratio]{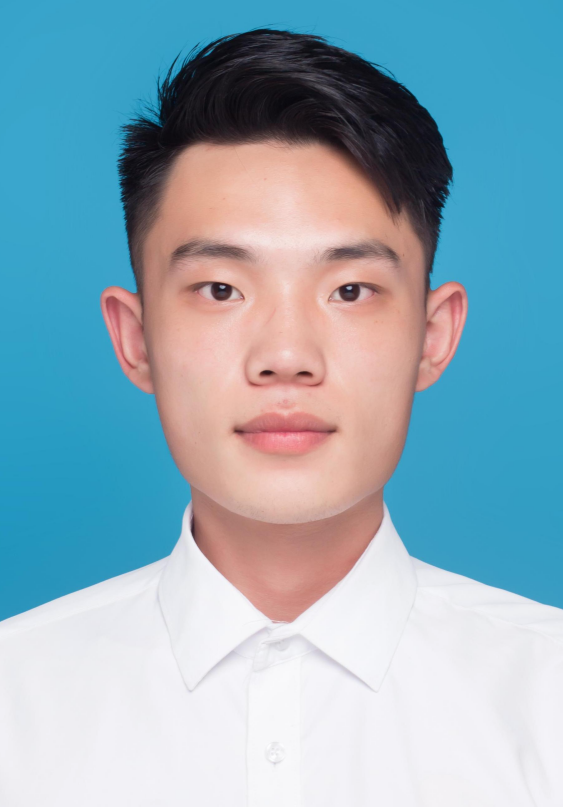}}] {Qinghui Chen} received the master’s degree in Qilu University of Technology. He is currently engaged in doctoral studies in Pattern Recognition and Intelligent Systems from Shandong University. His research focuses on time series forecasting,  spatiotemporal prediction, pattern recognition, and computer vision.
\end{IEEEbiography}

\vspace{-1cm}
\begin{IEEEbiography}
[{\includegraphics[width=1in,height=1.25in,clip,keepaspectratio]{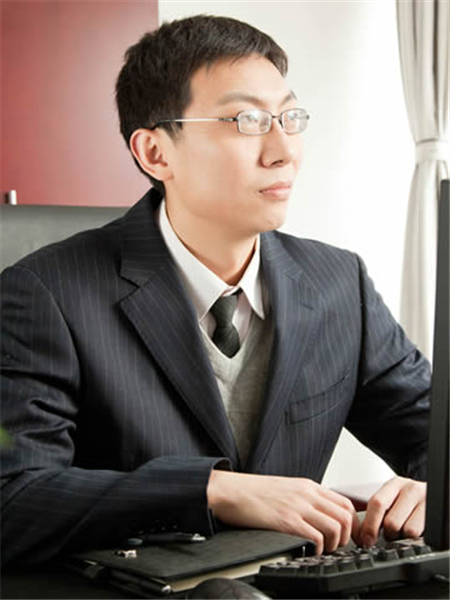}}] {Gang Li} received the Ph.D. degree in Management Science and Engineering from Harbin Institute of Technology, Harbin,  China. He is currently a Full Professor of the School of Computer  Science and Technology, Qilu University of Technology (Shandong  Academy  of  Sciences).  His current research interests include machine vision, pattern recognition.
\end{IEEEbiography}

\vspace{-1cm}
\begin{IEEEbiography}
[{\includegraphics[width=1in,height=1.25in,clip,keepaspectratio]{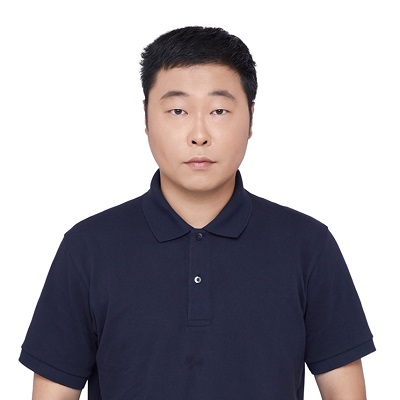}}] {Da Chen} received his Ph.D degree in applied mathematics from CEREMADE, University Paris Dauphine, PSL Research University, Paris, France, in 2017. From 2017 to 2019, he worked as a post-doctoral researcher at CEREMADE, University Paris Dauphine, and also at Centre Hospitalier National d'Ophtalmologie des Quinze-Vingts, Paris, France. Now he is working at CEREMADE, Paris, France. His research interests include variational methods, machine learning, minimal paths, and geometric methods with applications in image analysis and robotics.
\end{IEEEbiography}

\vspace{-1cm}
\begin{IEEEbiography}
[{\includegraphics[width=1in,height=1.25in,clip,keepaspectratio]{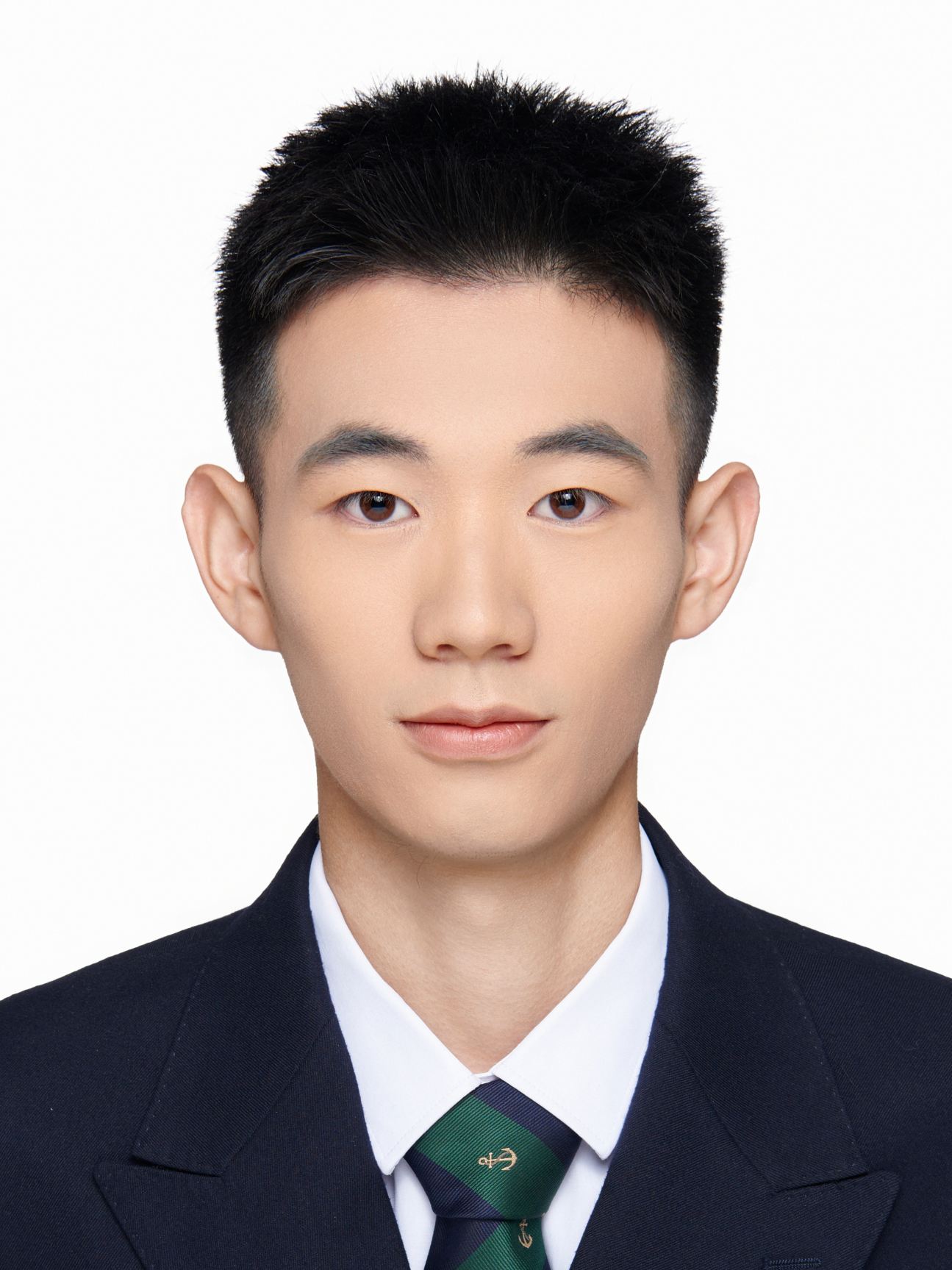}}] {Shuainan Jing} is currently pursuing a master's degree in Qilu University of Technology. His research interests include computer vision and deep learning.
\end{IEEEbiography}

\vspace{-1cm}
\begin{IEEEbiography}
[{\includegraphics[width=1in,height=1.25in,clip,keepaspectratio]{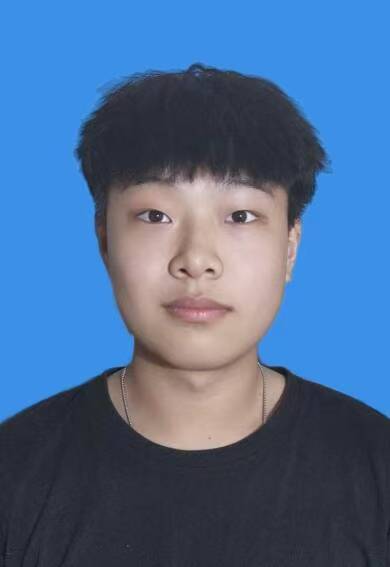}}] {He Wang} is currently pursuing a master's degree in Qilu University of Technology. His research interests include computer vision and deep learning.
\end{IEEEbiography}

\vspace{-1cm}
\begin{IEEEbiography}
[{\includegraphics[width=1in,height=1.2in,clip,keepaspectratio]{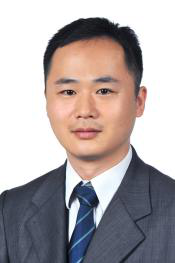}}] 
{Dagang Li} (Member, IEEE) received the Ph.D. degree in electrical engineering from Katholieke Universiteit Leuven, Leuven. He is currently an Associate Professor with the School of Computer Science and Engineering, Faculty of Innovation Engineering, Macau University of Science and Technology. His research interests include reinforcement learning, autonomous driving.
\end{IEEEbiography}

\vspace{-1cm}
\begin{IEEEbiography}
[{\includegraphics[width=1in,height=1.25in,clip,keepaspectratio]{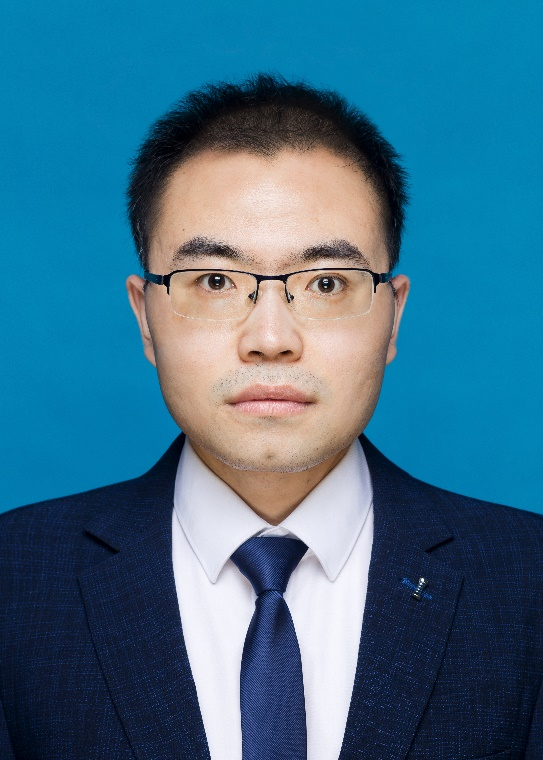}}]
{Cong Liu} received the BS and the MS degree in computer software and theory from Shandong University of Science and Technology, Qingdao, China, in 2013 and 2015 respectively. He received the PhD degree in the Department of Mathematics and Computer Science, Eindhoven University of Technology, 2019. He is an invited full professor in the NOVA Information Management School, Nova University of Lisbon. His research interests are in the areas of process mining, business process management, and artificial intelligence.
\end{IEEEbiography}

\vspace{-1cm}
\begin{IEEEbiography}
[{\includegraphics[width=1in,height=1.25in,clip,keepaspectratio]{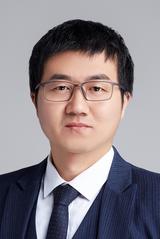}}]
{Cong Bai} (Member, IEEE) received the Ph.D. degree in signal and image processing from the National Institute of Applied Sciences, Rennes, France, in 2013. He is a Professor with the College of Computer Science and Technology, Zhejiang University of Technology, Hangzhou, China. His research interests include computer vision and multimedia processing.
\end{IEEEbiography}

\vspace{-1cm}
\begin{IEEEbiography}
[{\includegraphics[width=1in,height=1.25in,clip,keepaspectratio]{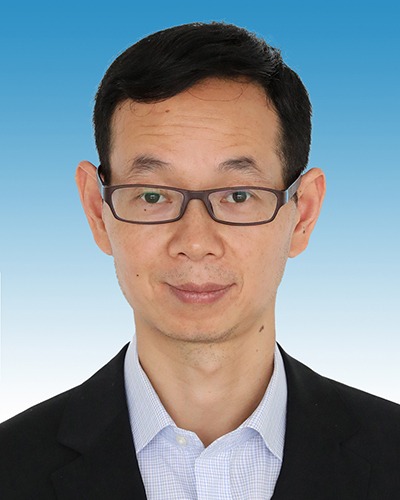}}]
{Shengyong Chen} (Senior Member, IEEE) received  the Ph.D. degree in computer vision from the City University of Hong Kong, Hong Kong, in 2003. He worked with the University of Hamburg from 2006 to 2007. He is currently a Professor with the Tianjin University of Technology, China. His research interests include computer vision, robotics, and image analysis. He is also a Senior Member of CCF and a fellow of IET. He received the Fellowship from the Alexander von Humboldt Foundation of Germany. He also received the National Outstanding Youth Foundation Award of China in 2013.
\end{IEEEbiography}

\vfill

\end{document}